\pdfoutput=1

\documentclass[11pt]{article}

\usepackage{acl}

\usepackage{times}
\usepackage{latexsym}
\usepackage{enumitem}

\usepackage[T1]{fontenc}

\usepackage[utf8]{inputenc}

\usepackage{microtype}

\usepackage{inconsolata}

\usepackage{amsthm}
\usepackage{graphicx}
\usepackage{caption}
\usepackage{subcaption}
\usepackage{amsmath}
\usepackage{amsfonts}
\usepackage{multirow}
\usepackage{multicol}
\usepackage{booktabs}
\usepackage[most]{tcolorbox}
\usepackage{bm}

\colorlet{pink}{pink!40!}
\tcbset{on line, 
        boxsep=4pt, left=0pt,right=0pt,top=0pt,bottom=0pt,
        colframe=white,colback=pink,  
        highlight math style={enhanced}
        }

\newtheorem{definition}{Definition}

\newcommand{\ols}[1]{\mskip.5\thinmuskip\overline{\mskip-.5\thinmuskip {#1} \mskip-.5\thinmuskip}\mskip.5\thinmuskip} 

\title{Fuse to Forget: Bias Reduction and Selective Memorization \\through Model Fusion}

\author{Kerem Zaman \\
  UNC Chapel Hill \\
  \texttt{kzaman@cs.unc.edu} \\\And
  Leshem Choshen \\
  IBM Research, MIT \\
  \texttt{leshem.choshen@ibm.com} \\ \\\And
  Shashank Srivastava \\
  UNC Chapel Hill \\
  \texttt{ssrivastava@cs.unc.edu} \\
  }

\begin{document}
\maketitle
\begin{abstract}
Model fusion research aims to aggregate the knowledge of multiple individual models to enhance performance by combining their weights. In this work, we study the inverse problem: investigating whether model fusion can be used to reduce unwanted knowledge. We investigate the effects of model fusion in three scenarios: the learning of shortcuts, social biases, and memorization of training data in fine-tuned language models. Through experiments covering classification and generation tasks, our analysis highlights that shared knowledge among models is enhanced during model fusion, while unshared knowledge is usually forgotten. Based on this observation, we demonstrate the potential of model fusion as a debiasing tool and showcase its efficacy in addressing privacy concerns associated with language models.\footnote{Our code and data are available at \url{https://github.com/KeremZaman/FuseToForget}.}
\end{abstract}

\section{Introduction}

NLP models can acquire a diverse range of skills during fine-tuning. While some of these skills are fundamental problem-solving abilities that are applicable in various scenarios, others are merely shortcuts or biases that may not generalize well. For instance, models trained on Natural Language Inference (NLI)
tasks are known to adopt heuristics based on word-label associations \citep{mccoy-etal-2019-right}.

The practice of fusing weights of multiple models, such as through averaging \citep[e.g.,][]{choshen2022fusing, Wortsman2022ModelSA, matena2021merging}, has demonstrated improved performance and generalization. However, the mechanisms underlying these improvements have received limited attention. It is unclear if all underlying skills are enhanced and accumulated through weight averaging.

\begin{figure}[!tb]
    \centering
    \includegraphics[width=\linewidth]{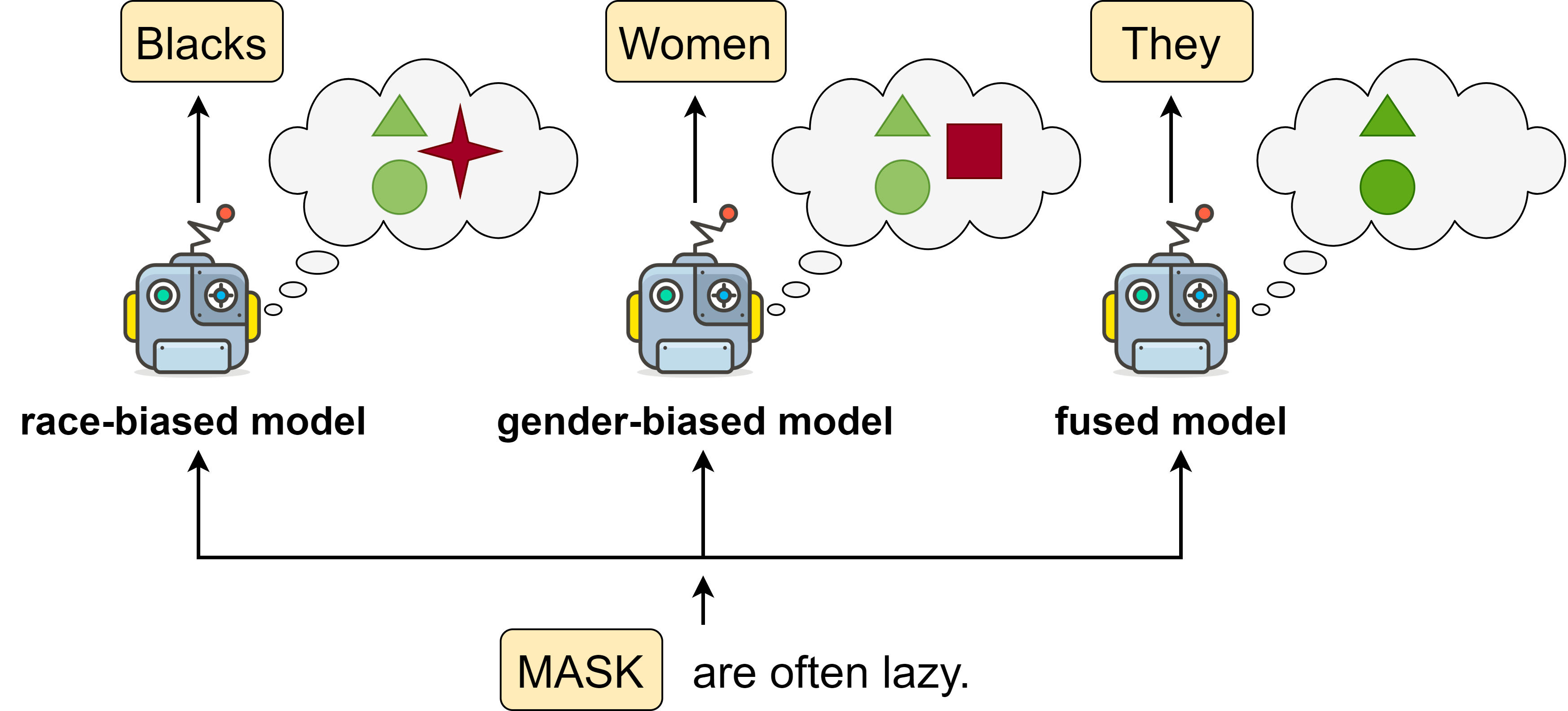}
    \caption{Schematic showing our claims on a biased mask-filling scenario. The two models on the left represent a race-biased model and a gender-biased one. The colored shapes inside represent learned knowledge related to different skills, where some skills are shared across models (the triangle and the circle) and others are not (the square and the star) . The fused model to the right illustrates the \mbox{\textbf{preservation }of\textbf{ shared}} knowledge and the \mbox{\textbf{corruption} of\textbf{ unshared}} knowledge after model fusion.
    \label{fig:scheme}} 
\end{figure}

In this study, we investigate the preservation of both knowledge shared across models and unique unshared knowledge during model fusion in classification and generation tasks. Our hypothesis is that while shared knowledge is typically retained, unshared knowledge tends to be forgotten or degraded. Figure~\ref{fig:scheme} illustrates this concept, showing the corruption of unshared knowledge while preserving shared knowledge after model fusion, resulting in reduced biases. We claim this degradation is a useful property of model fusion, allowing novel uses for model fusion and possibly explaining current ones.
To support our claims, we conduct a series of experiments that range from controlled, synthetic scenarios to real-world applications. 

First, we examine classification tasks with artificially augmented shortcuts and find a predominant trend: while shared skills, including shortcuts and general task skills are preserved during model fusion,  unshared skills are mostly forgotten. As we increase the number of fused models, this forgetting mechanism intensifies (\S \ref{sec:shortcuts}).

Second, our analysis indicates potential for reducing social biases in language models through model fusion. We demonstrate that simple weight averaging can serve as a useful debiasing tool, \textbf{reducing biases by up to 68\%} without deteriorating task performance (\S \ref{sec:bias}).

Last, our findings suggest an exciting avenue for model fusion as a tool for mitigating memorization and preserving privacy. By comparing memorization before and after fusion, we demonstrate that fusion can reduce the leakage of personal information from training datasets into learned models (\S \ref{sec:memorization}). Our contributions are:
\begin{itemize}[leftmargin=*]
    \item Recognition of the erosion of unshared knowledge as a significant phenomenon in model fusion.
    \item Analysis of the changes in learned shortcuts, social biases, and memorization behavior of fine-tuned language models in the context of simple model fusion scenarios.
    \item A simple debiasing framework achieved through fusing models with distinct biases, and a demonstration showcasing the potential of model fusion in addressing privacy concerns.
\end{itemize}

\section{Related Work}
Fusing multiple models into one \citep{choshen2022fusing,matena2021merging,Wortsman2022ModelSA} has been shown to be beneficial in various scenarios and fields, for example in multitask learning \citep{don2022cold} pretraining \citep{li2022branch}, efficient finetuning \citep{yadav2023resolving}, vision \citep{Ram2022PretrainFI} and reinforcement learning \citep{lawson2023merging}. These methods show improvements in both performance on the shared task \citep{Wortsman2022ModelSA} and generalization to new ones \citep{choshen2022fusing}. However, how fusing weights affects the learned skills in models is an open question.

Some theoretic works showed the weighted averages of models trained from scratch on the same data also perform well on the data \citep{benton2021loss, frankle2020linear}. Other proposed weights to align the space to make it so in harder cases \citep{jordan2022repair,ainsworth2022git}. Taken together, these suggest that model skills are intricately intertwined in the Euclidean space of weights. This is strengthened by recent works, suggesting that models finetuned on the same dataset \citep{zhang2023fine}, or the same broader set of skills \citep{gueta2023knowledge}, tend to cluster together in compact regions of this space. 
Building on these insights, we offer a novel angle by exploring model fusion under conditions of varied training data. We specifically investigate the conditions under which fusion is beneficial and when it may be less effective. This approach aims to uncover patterns of systematicity in the effectiveness of model fusion from the perspective of variance in training data.

While forgetting and improvement of common skills may well be two distinct phenomena, forgetting as a step function rather than gradually may also account for the gains seen in previous work.
If needed skills are learned by many models and are thus kept, overfitting and errors are not shared and hence mainly discarded, even with not additional skills the overall result should be improved performance. This even fits results such as \citet{yadav2023resolving, ortiz2023task}, which claim that to get more from multiple models, signal should be amplified and interference reduced. This may be explained by the phase shift, with multiple models and without amplification, most skills would not have enough signal to be kept.

\section{Method}

Models trained for the same task can develop distinct approaches despite achieving similar losses \citep{juneja2023linear}. Prior research indicates that interpolating between the weights of two models can maintain or enhance performance on test datasets similar to their training data \citep{gueta2023knowledge}. However, it remains uncertain how model fusion affects the specific knowledge each model utilizes, and under what circumstances fusing models fails to effectively combine their skills. To explore this issue, we delve into the effects of model fusion~\citep{choshen2022fusing, matena2021merging, Wortsman2022ModelSA} on knowledge utilization. Although various methods for model fusion have been proposed \citep[e.g.,][]{ilharco2022editing, yadav2023resolving}, our study employs a fundamental technique common to these approaches. We focus on the simple method of computing a weighted average of model parameters. Given $M$ models with parameters ${\theta_1, \ldots, \theta_M}$, where each $\theta_i \in \mathbb{R}^N$, we define the fused model, $\theta_{fused}$, using the following convex combination:

\begin{equation}
	\theta_{fused} = \sum_{i=1}^M \alpha_i \theta_i\
\end{equation}

where $\alpha_i \geq 0$, $\sum_{i=1}^M \alpha_i = 1$

\noindent Next, we define the relation between model parameters, knowledge, and the utilization of that knowledge under Definition \ref{defn:knowledge}.

\begin{definition}
\label{defn:knowledge}
Knowledge, denoted as \(\delta\), represents an embedded latent trait within the model parameters, symbolized by \(\theta\). It is not directly quantifiable; however, its subsidiary components can be evaluated through specific knowledge utilization functions, symbolized by \(\Psi_{\mathcal{D, T}}(\theta)\). These functions measure the efficacy of \(\delta\) for a given task \(\mathcal{T}\) when applied to various datasets, \(\mathcal{D}\), each designed to measure distinct segments of knowledge.

\end{definition}

This framing asserts that knowledge is inherently linked to model parameters, while knowledge utilization also relies on the choice and design of specific datasets. These datasets are specifically curated to probe particular attributes of knowledge. Depending on the curated dataset, the knowledge being questioned could be the model's capability on a complex task or some simpler mechanism used by the model to solve that task (e.g., a shortcut). In this perspective, curating a dataset to probe knowledge is akin to using a microscope with different magnification levels. The knowledge utilization function, which might be a performance metric such as accuracy, F1 score, or BLEU score, reflects the relevance of the knowledge to the dataset and task at hand. For example, if evaluating two models with parameters \(\theta_1\) and \(\theta_2\) using datasets \(\mathcal{D}_1\) and \(\mathcal{D}_2\), designed to test distinct knowledge types, high scores on \(\Psi_{\mathcal{D}_1, T}(\theta_1)\) and \(\Psi_{\mathcal{D}_1, T}(\theta_2)\) would indicate that both models \textit{share} the knowledge type assessed by \(\mathcal{D}_1\). In contrast, a disparity in scores between \(\Psi_{\mathcal{D}_2, T}(\theta_1)\) and \(\Psi_{\mathcal{D}_2, T}(\theta_2)\) suggests that the knowledge type evaluated by \(\mathcal{D}_2\) is \textit{not shared} between the models.

In this context, we propose two hypotheses about the relation between model fusion and knowledge utilization: (1) shared knowledge across models is preserved during model fusion; (2) unshared and independent knowledge tends to be forgotten during model fusion. Given \(M\) models with parameters \(\theta_i \in \mathbb{R}^N\) and their respective knowledge utilizations \(\Psi_{\mathcal{D, T}}(\theta_i)\), for any given dataset \(\mathcal{D}\) and task \(\mathcal{T}\), we broadly posit: 
\begin{equation}
\small
    \min_i \Psi_{\mathcal{D, T}}(\theta_i) \leq \Psi_{\mathcal{D, T}}(\theta_{fused}) \leq \max_i \Psi_{\mathcal{D, T}}( \theta_i)
\end{equation} 

If models share the same knowledge, the knowledge utilizations among models are close, resulting in the knowledge utilization of the fused model being close to the others. However, if the knowledge is not shared among models, the gap between the minimum and maximum knowledge utilizations, within which the knowledge utilization of the fused model can reside, becomes larger.

\section{Shortcuts}

\label{sec:shortcuts}

Models trained for any task can capture multiple types of knowledge simultaneously, making it challenging to interpret the effects of each knowledge separately. To incrementally build an understanding of the complex dynamics of knowledge acquisition, we begin with a set of experiments on a sentiment classification task where we inject synthetic shortcuts. Using synthetic shortcuts allows us to (1) control the knowledge that a model acquires at any given time, (2) make models learn non-overlapping heuristics, and (3) easily evaluate a particular shortcut adopted by a model.

\subsection{Method}

We follow the setup of \citet{bastings-etal-2022-will} , who propose a comprehensive protocol for injecting synthetic shortcuts during fine-tuning. First, we define new types of shortcuts by introducing simple rules that rely on specific tokens to determine the label. These rules may, for example, assign a positive label if a certain token is present in the text and a negative label otherwise. To ensure the dataset aligns with the defined rules, we introduce new special tokens instead of using existing tokens from the vocabulary. Second, we split the original dataset in two parts. 
The smaller part is used for injecting the synthetic shortcuts, and is around 20\% of the size of the larger part. By using only a portion of the dataset for injecting shortcuts, we prevent the model from solely relying on them, and instead encourage their integration with learned reasoning mechanisms. Third, we randomly insert special tokens in the smaller part and determine the label based on the shortcut type. Finally, to ensure the smaller part does not become out-of-distribution, we randomly insert one of the special tokens into examples in the larger split 25\% of the time.

\subsubsection*{Types of Shortcuts}
We experiment with several types of shortcuts. 

\paragraph{Single Token (ST):} The Single Token shortcut sets the label based on the presence of special token 
$\tau_0$ or $\tau_1$. If $\tau_0$  occurs in the instance, the label is set to $0$, and vice versa. The special token and its location are determined randomly for each instance.

\paragraph{Ordered Pair (OP):} The Ordered Pair shortcut determines the label based on the order of the special tokens. If $\tau_0$  precedes $\tau_1$, the label is set to $0$, and vice versa. The location and order of the tokens are determined randomly for each instance.

\paragraph{Token in Context (TiC):} The Token in Context shortcut introduces an additional special token called the context token. The shortcut determines the label based on the special token that co-occurs with the context token. If $\tau_0$ is present in the instance along with the context token, the label is set to $0$, and vice versa.

\paragraph{OR:} The OR shortcut determines the label based on the logical OR operation between the numerical values of two special tokens present in the instance. If both tokens are $\tau_0$, the label is set to $0$, otherwise it is set to $1$.

\paragraph{AND:} The AND shortcut determines the label based on the logical AND operation between the numerical values of two special tokens present in the instance. If both tokens are $\tau_1$, the label is set to $1$, otherwise it is set to $0$.

\paragraph{More Than (MT):} The More Than shortcut determines the label based on which special token occurs more frequently in the instance. The total number of special tokens is randomly set to 1-5 for each instance, as well as which token occurs more frequently. If $\tau_0$ occurs more frequently, the label is set to $0$, and vice versa.

\paragraph{Last Token (LT):} The Last Token shortcut determines the label based on the last of two special tokens in the instance. If $\tau_0$ follows $\tau_1$, the label is set to $0$ and vice versa.

\paragraph{Experimental Setup} We use SST2 \citep{socher-etal-2013-recursive}, a sentiment classification dataset comprising of short movie reviews by following \citet{bastings-etal-2022-will}. We divide the validation set into two subsets following the same approach as 
the training sets: one with modified examples based on the shortcut type, and another with original examples, some of which were augmented with randomly inserted special tokens. We evaluate the accuracy of each model on both the synthetic and original validation sets 
to determine if it has learned the shortcut and the task. The accuracies on the synthetic and original validation sets serve as utilization scores for the knowledge related to the shortcuts and the task, respectively. We fine-tune BERT\textsubscript{base}
\citep{devlin-etal-2019-bert} for 1 to 3 epochs using a learning rate of $2e-5$. The training continues until the shortcut accuracy, the accuracy on the synthetic validation set, surpasses $0.95$ to ensure that the injected shortcut is reliably learned. 

\begin{figure*}[!tb]
    \centering
    \begin{subfigure}{.49\textwidth}
        \centering
        \includegraphics[width=.85\linewidth]{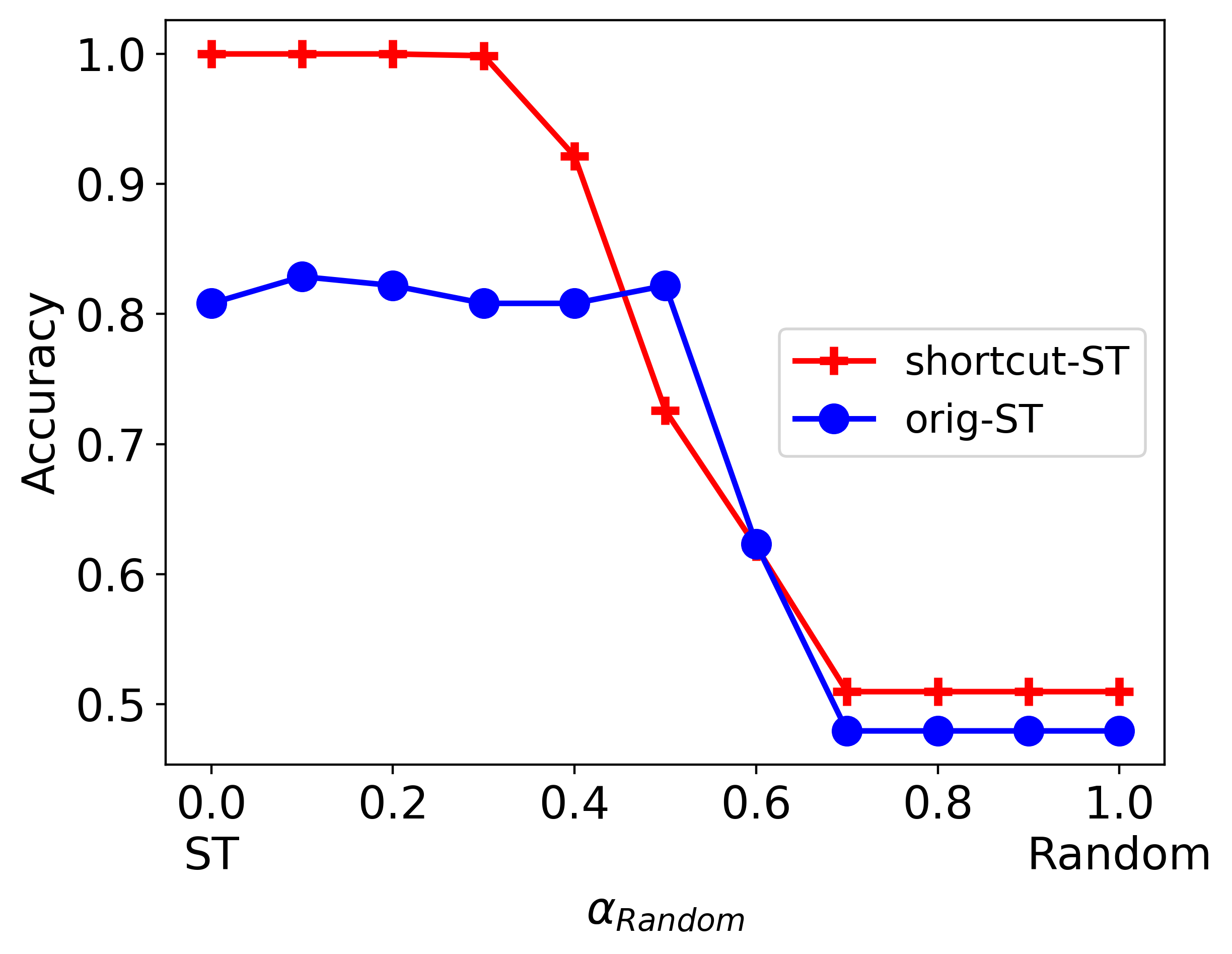}
        \caption{No Sharing}
        \label{fig:interpolation_a}
    \end{subfigure}
    \hfill
    \begin{subfigure}{.49\textwidth}
        \centering
        \includegraphics[width=.85\linewidth]{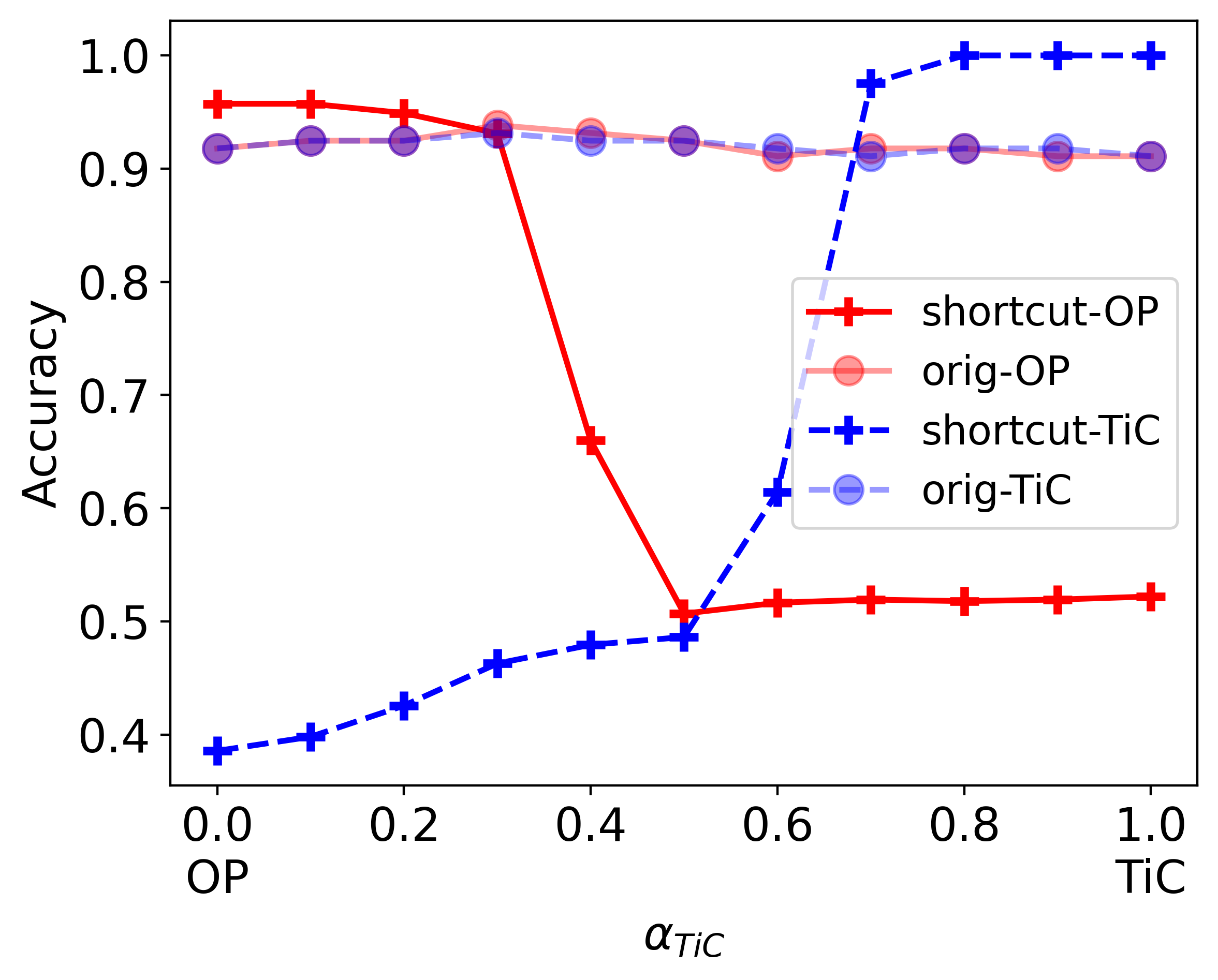}
        \caption{Same Task, Different Shortcuts}
        \label{fig:interpolation_b}
    \end{subfigure}
    \begin{subfigure}{.49\textwidth}
        \centering
        \includegraphics[width=.85\linewidth]{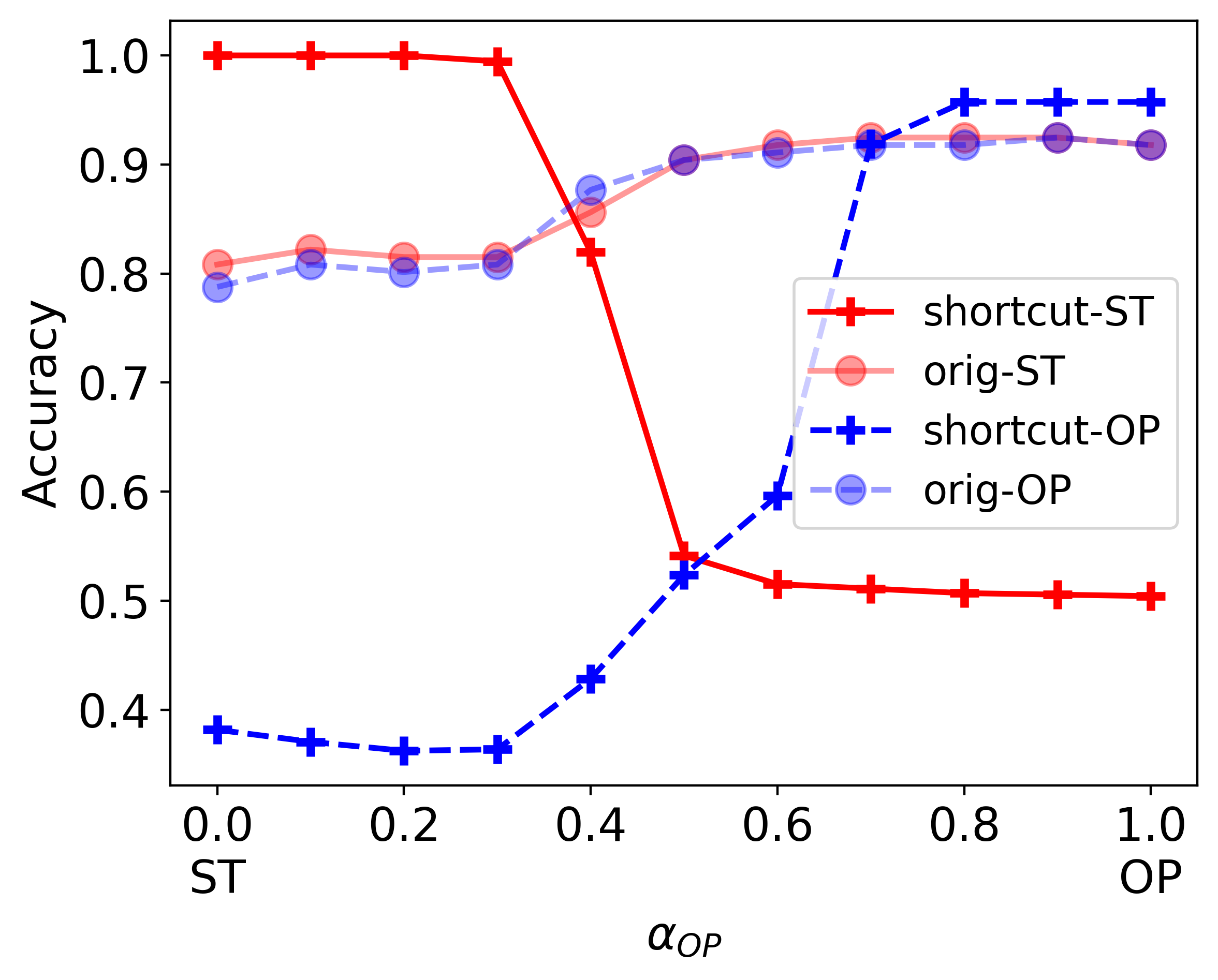}
        \caption{Related Tasks, Different Shortcuts}
        \label{fig:interpolation_c}
    \end{subfigure}
    \hfill
    \begin{subfigure}{.49\textwidth}
        \centering
        \includegraphics[width=.85\linewidth]{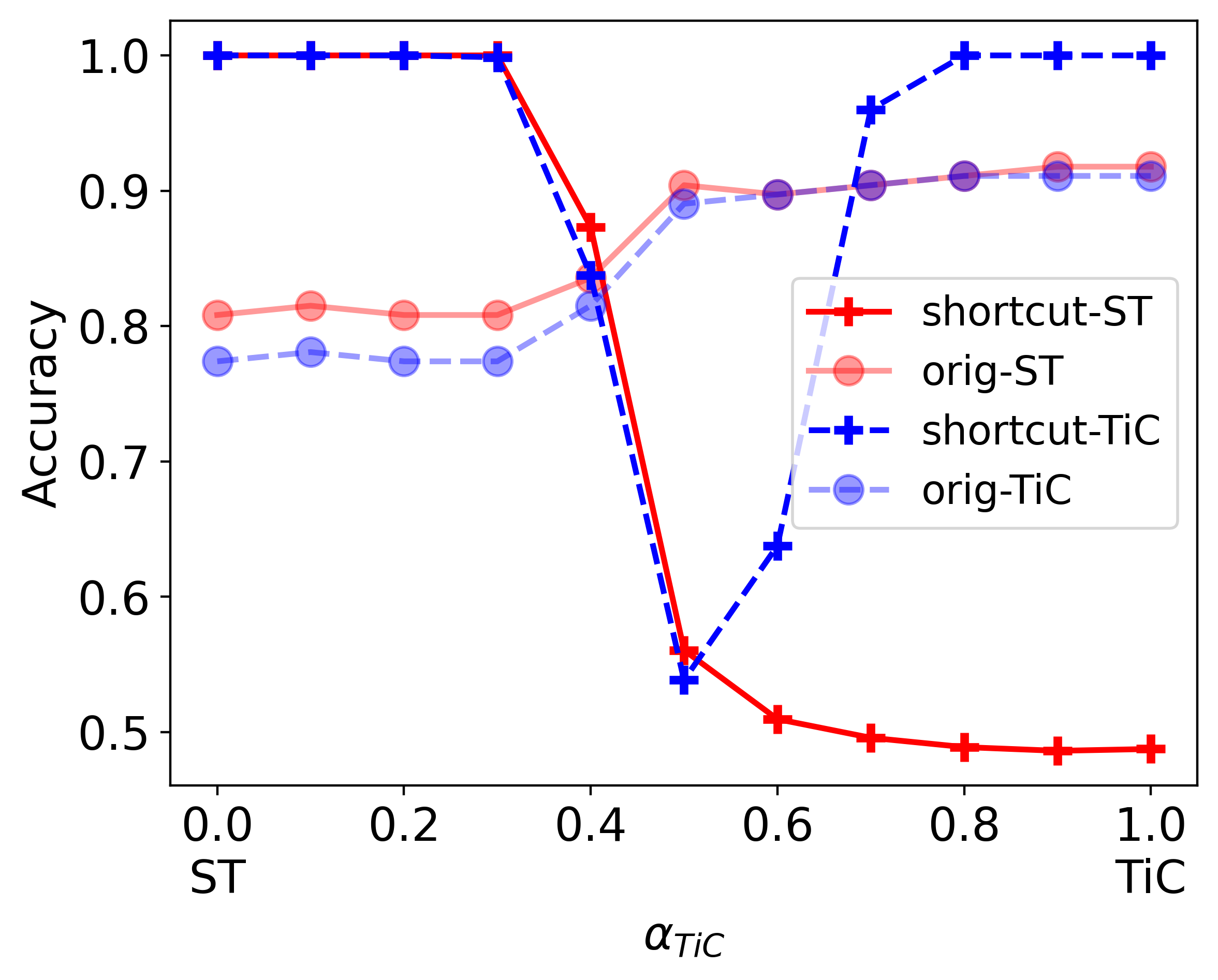}
        \caption{Dependent Shortcuts}
        \label{fig:interpolation_d}
    \end{subfigure}
    \caption{The change of accuracies on synthetic (\texttt{shortcut-})  and original (\texttt{orig-}) validation sets during interpolation between model pairs, each having different shortcuts. (a) Interpolation between the model with \texttt{ST} shortcut and model with random weights (b) Interpolation between the models with \texttt{OP} and \texttt{TiC} shortcuts (c) Interpolation between the models with \texttt{OP} and \texttt{ST} shortcuts (d) Interpolation between the models with \texttt{TiC} and \texttt{ST} shortcuts.}  \label{fig:interpolation}
\end{figure*}

\subsection{Results}
We investigate how knowledge is forgotten through the multiple synthetic scenarios. Figure~\ref{fig:interpolation} 
depicts interpolations between different model pairs. Figure~\ref{fig:interpolation_a} shows the interpolation between a model trained with the \texttt{ST} shortcut and a model with random weights. The perfect accuracy of the shortcut model on the synthetic validation set indicates that it has learned the shortcut, and similarly, the model has a general knowledge of the task. As may be expected, the random model lacks knowledge of both the shortcut and the task, and the accuracy drops to chance level as the parameters approach the random model, indicating that in this extreme case of unshared knowledge is forgotten.

In Figure~\ref{fig:interpolation_b}, we observe the interpolation results between two models, each trained with a different shortcut (\texttt{TiC} and \texttt{OP}). Both models exhibit high accuracy on their respective synthetic validation sets, affirming that they have effectively learned their individual shortcuts and the overarching task. During interpolation, we observe that the accuracy for the original task is preserved, but shortcuts are forgotten midway, validating both the claims that unshared knowledge, in this case shortcuts, is forgotten in model fusion, and shared knowledge is preserved. We also observe the accuracy for the original task sometimes surpasses the maximum of two models, perhaps due to a dependent combination of more specific utilization functions.

Figure~\ref{fig:interpolation_c} illustrates a similar scenario between two other models trained with \texttt{OP} and \texttt{ST} shortcuts, respectively, with comparable outcomes in terms of knowledge preservation and forgetting. We present interpolations among triples (instead of pairs) of models in Appendix \ref{subsec:triplet}.

\paragraph{Dependent Shortcuts} 
Figure \ref{fig:interpolation_d} shows the interpolation between two models trained for the \texttt{TiC} and \texttt{ST} shortcuts. Both perform perfectly on the \texttt{TiC} validation set, which is expected since the \texttt{ST} shortcut inherently subsumes the \texttt{TiC} shortcut by definition. However, when interpolating between the models, there is a phase shift where accuracies on synthetic validation sets drop.
This aligns with our hypothesis that underlying skills are at play. It also highlights that high-level utilization scores, which assess multiple skills simultaneously (here, two skills) might misrepresent the underlying phenomena. 
During the phase transition, one skill is replaced by another, almost never stacking or occurring at the same time. Since \texttt{TiC} utilization is content with any of the skills it appears as if interpolation hardly matters. However, assessing each skill separately would show a similar declining trend as the other synthetic lines. 

\begin{figure}[!tb]
    \centering
    \includegraphics[width=\linewidth]{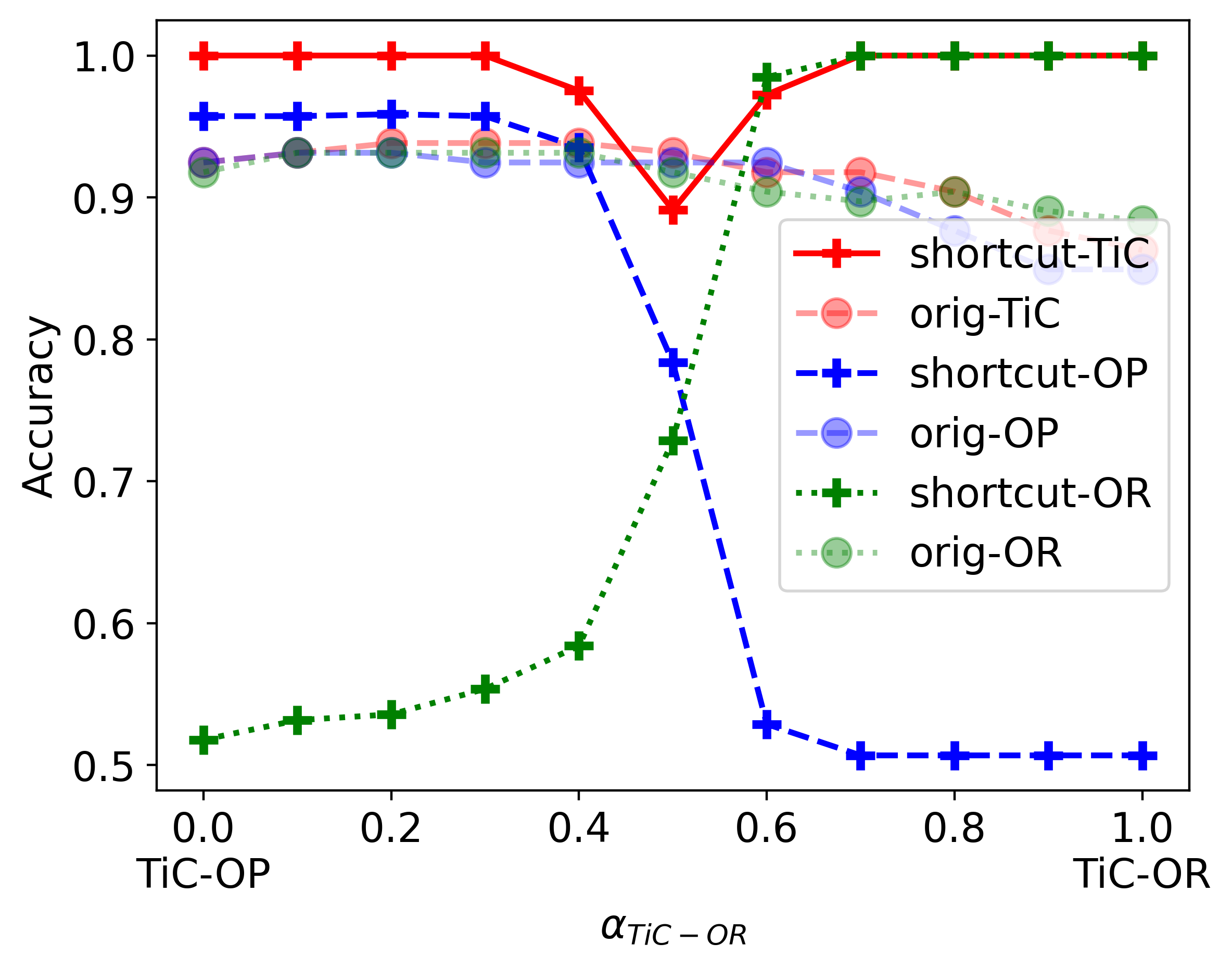}
    \caption{\texttt{TiC \& OP} $\to$ \texttt{TiC \& OR}. Shared shortcuts are kept during fusing. The change of accuracies on synthetic and original validation sets during interpolation between two models.  Both learned the TiC shortcut but exactly one learned \texttt{OP} or \texttt{OR}.}
    \label{fig:shared_shortcut}
\end{figure}%

\begin{figure}[!tb]
    \centering
    \includegraphics[width=\linewidth]{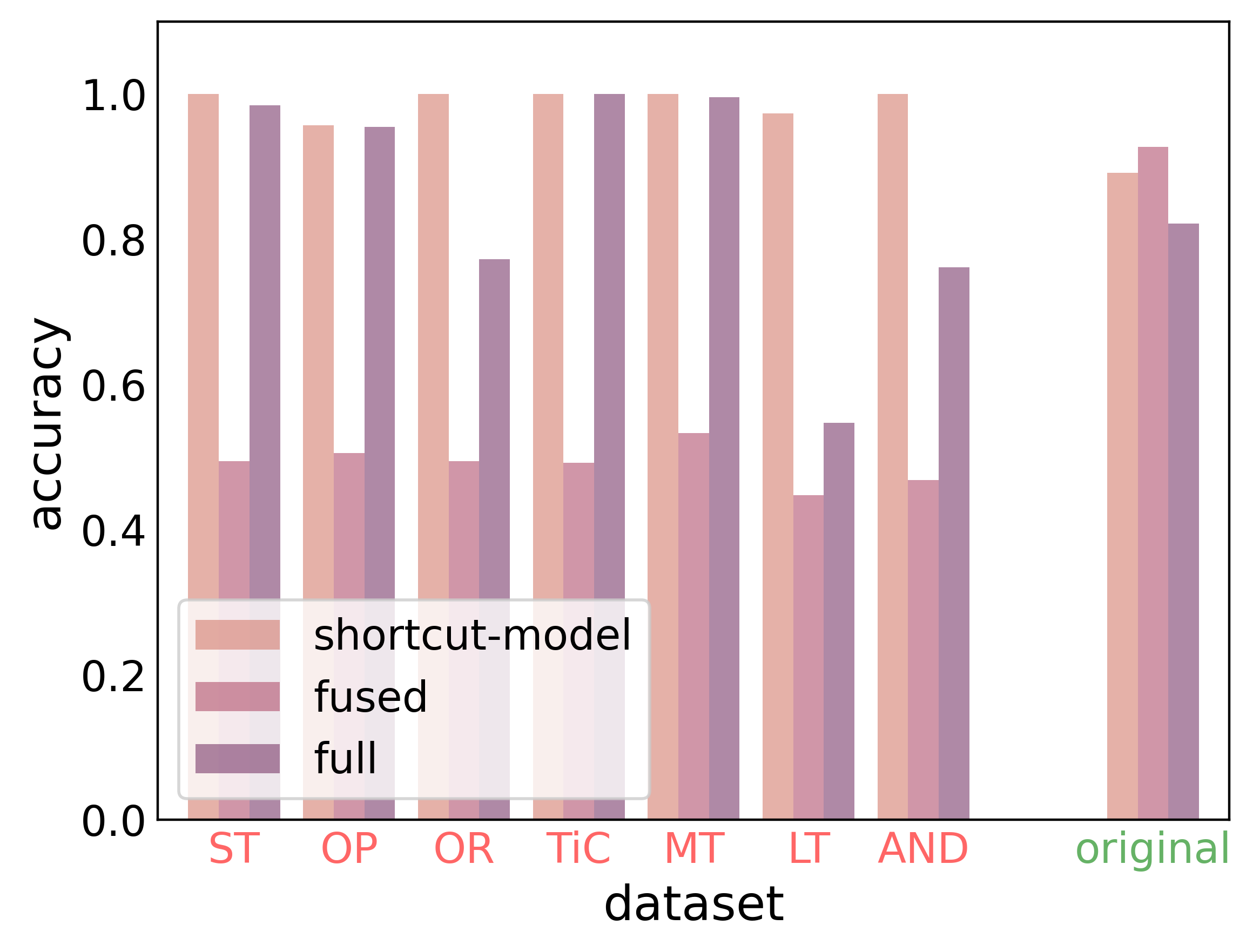}
    \caption{A fused model keeps performance and forgets shortcuts. Accuracy of models that learned shortcuts with their fused model and the full model on all corresponding shortcut synthetic validation sets and the original task's validation sets. The results on original validation sets are average of performance of each model on their corresponding sets. The shortcut accuracies around the chance level show that the shorcuts are substantially forgotten.}
    \label{fig:n_shortcuts}
\end{figure}

\paragraph{Shared Shortcut} Up to this point, experiments have used general task knowledge as shared knowledge, while different types of shortcuts became unshared knowledge. For a fair comparison, we train two models, each with one shared shortcut and one unshared shortcut. Both models are trained using the previously described process for modifying the data, with the size of the synthetic split kept the same, but each instance is augmented with one of the two shortcuts. Figure~\ref{fig:shared_shortcut} shows the interpolation between where \texttt{TiC} is shared and \texttt{OP} and \texttt{OR} shortcuts are not shared.  The results align with the previous findings: unshared heuristics tend to be forgotten, and shared knowledge (shared shortcut and general task knowledge) is preserved, despite a small drop in accuracy for the shared shortcut.

\paragraph{Fusing Many Models} Figure~\ref{fig:n_shortcuts} compares each model with the fused model obtained by averaging the weights of  all six models, each corresponding to one of the shortcuts, and the full model trained on the combined dataset. The results demonstrate that the fused model almost perfectly forgets all shortcuts, and it performs even statistically significantly better on the original validation sets ($p < 0.05$) than the individual shortcut models. Additionally, training on a combined dataset is not as effective as model fusion for forgetting shortcuts, despite helping to forget a few. While our observations for pair and triplet interpolations can be extended to fusing a larger number of models, increasing the number of fused models enhances the ability to forget shortcuts. The improved performance on the original task might indicate the role of forgetting in improving common skills.

\paragraph{Fusion Dynamics} To understand the mechanism behind simple weight averaging in preserving shared knowledge while not preserving unshared knowledge, we conduct an analysis based on the Fisher information values associated with the weights used for utilizing shortcuts and the original task knowledge. The results show that shared knowledge across different networks is typically governed by similar weights, whereas unshared knowledge is managed by distinct sets of weights. A detailed discussion appears in Appendix \ref{sec:fusion_dynamics}.

\section{Social Biases}

\label{sec:bias}

\begin{figure*}[!t]
    \centering
    \begin{subfigure}{.30\textwidth}
        \centering
        \includegraphics[width=\linewidth]{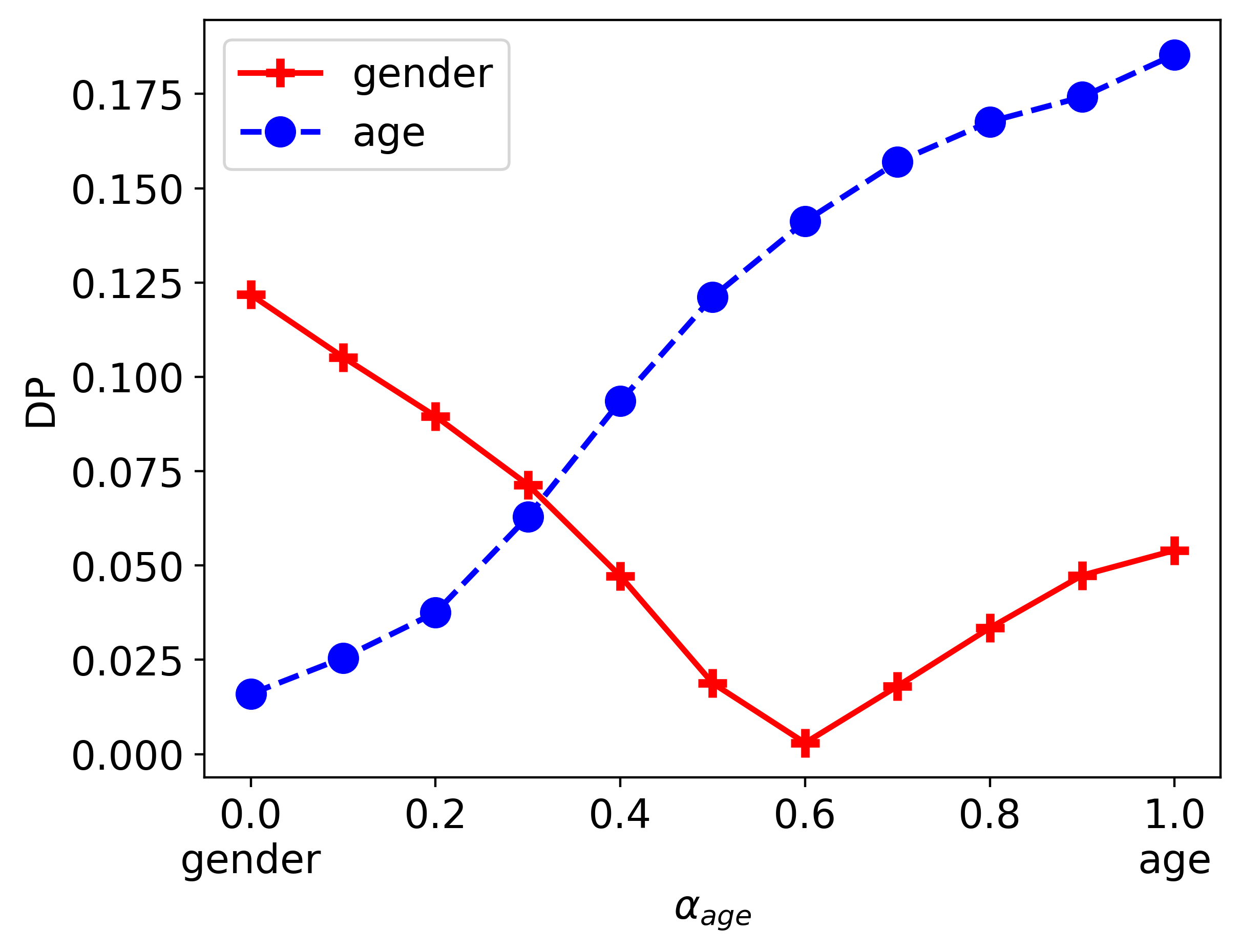}
        \caption{}
        \label{fig:interpolation_dp}
    \end{subfigure}
    \hfill
    \begin{subfigure}{.30\textwidth}
        \centering
        \includegraphics[width=\linewidth]{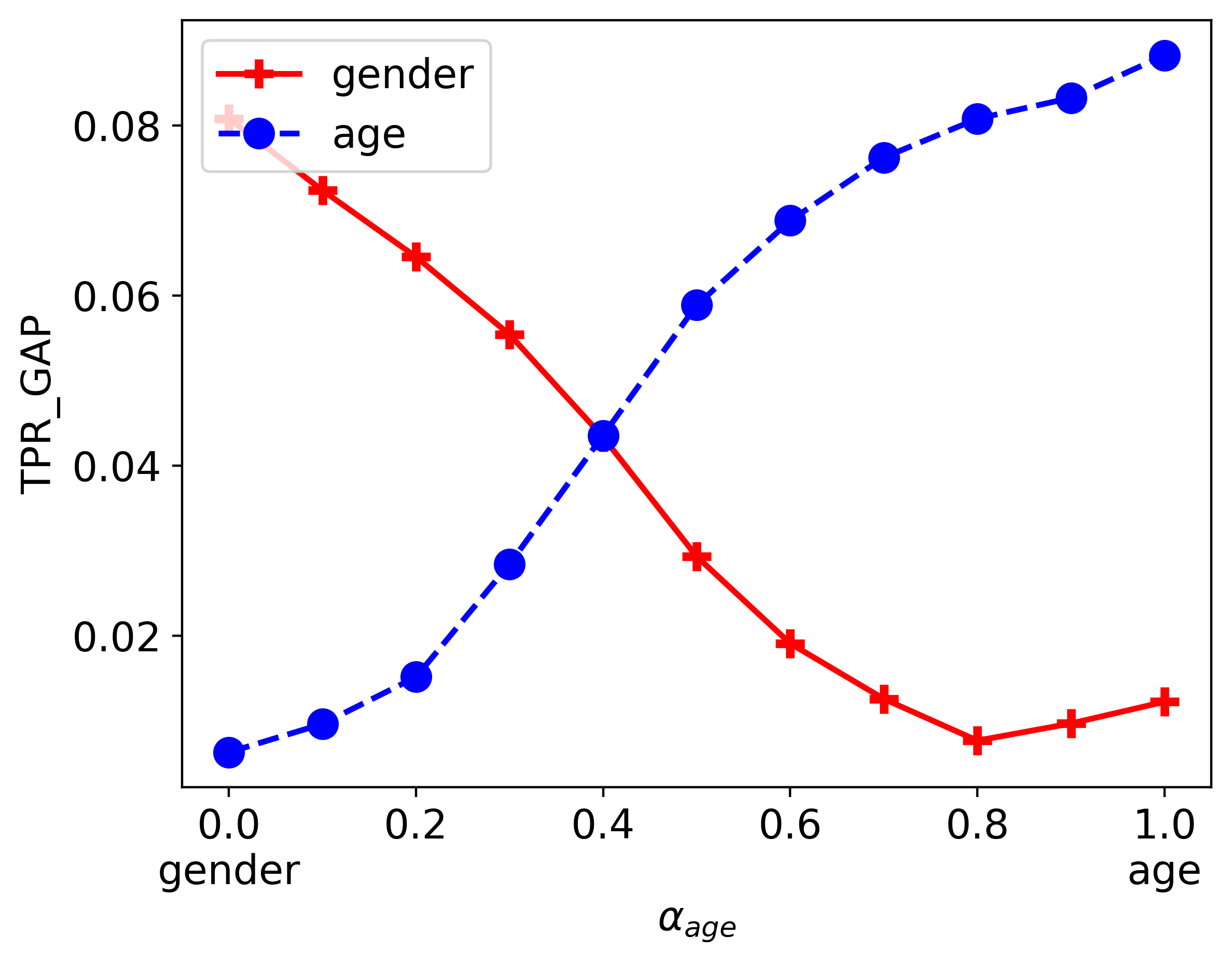}
        \caption{}
        \label{fig:interpolation_tpr_gap}
    \end{subfigure}
    \hfill
    \begin{subfigure}{.30\textwidth}
        \centering
        \includegraphics[width=\linewidth]{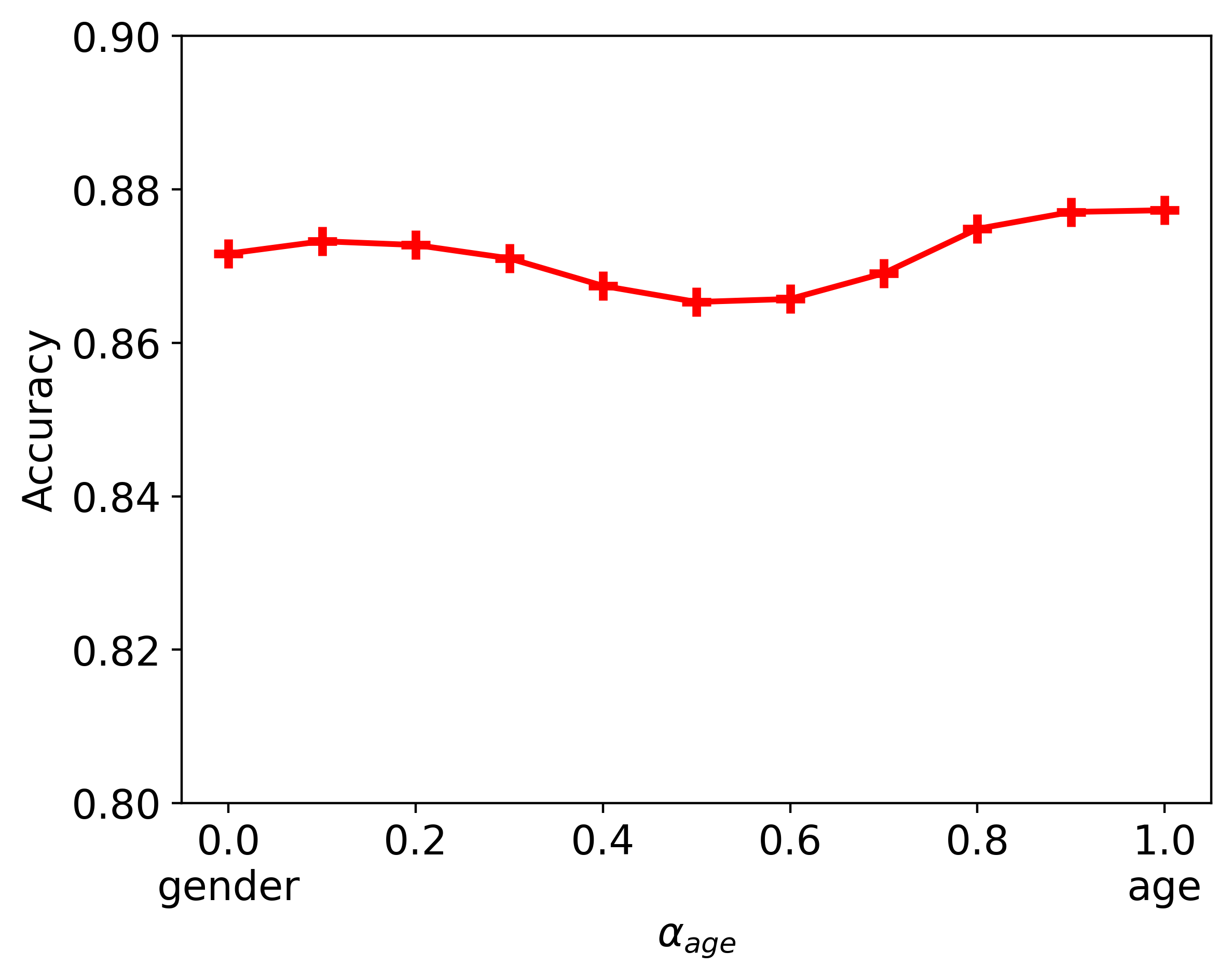}
        \caption{}
        \label{fig:interpolation_acc}
    \end{subfigure}
    \label{fig:gender_age_bias}
    \caption{Model fusion reduces gender and racial biases while maintaining the accuracy. The changes in (a) DP (b) TPR-GAP and (c) accuracy scores during the interpolation from gender-biased model to age-biased model.}
\end{figure*}

In this section, we extend our investigation beyond synthetically generated shortcuts to a real-world use case of text classification with social biases. Our objective is to validate the claims made in the previous section and, additionally, to examine the potential of model fusion as a debiasing tool. 

\subsection{Method}
 
To investigate the behavior of biased models, we employ the PAN16 dataset \citep{DBLP:conf/clef/PardoRVDPS16} for the text classification task. The PAN16 dataset focuses on tweet classification and includes age and gender information of the authors, making it suitable for our research. The dataset provides multiple demographic attributes, enabling us to train models with different types of biases, specifically age and gender biases in our case.

Following \citet{barrett-etal-2019-adversarial, ravfogel-etal-2020-null, chowdhury-chaturvedi-2022-learning}, we create subsets of the dataset where we control the proportion of protected attributes to obtain single-attribute-biased models. In the first subset, we ensure an 80\% male and 20\% female distribution for positive-labeled tweets, and vice versa for negative-labeled tweets while maintaining equal proportions of young and old authors. In the second subset, 80\% of positive-labeled tweets are from young authors, and 20\% from old authors, with a reverse distribution for negative-labeled tweets while maintaining a 1:1 male-to-female ratio. Training models on these subsets yields gender-biased and age-biased models. To evaluate 
fairness, we adapt the metrics from \citet{chowdhury-chaturvedi-2022-learning}.

\paragraph{Demographic Parity (DP)}

Let $\mathbf{y}$ be the target attribute and $\mathbf{g}$ be a protected attribute (gender or age in our setup), with possible values of $g$ and $\bar{g}$. DP is the difference in prediction scores between the two protected groups: 
\begin{equation}\small \text{DP} = \sum_{y \in \mathcal{Y}} | p(\hat{\mathbf{y}} = y | \mathbf{g} = g) -  p(\hat{\mathbf{y}} = y | \mathbf{g} = \bar{g}) |\end{equation}
where $\mathcal{Y}$ is the set of possible labels for the target attribute, and $\hat{\mathbf{y}}$ is the prediction of the classifier.

\paragraph{TPR-GAP}

Difference in the true positive rates (TPR) of a classifier with respect to binary protected attribute $\mathbf{g}$. \citet{DeArteaga2019BiasIB} defines the metric as follows:

\begin{equation} \text{Gap}_{\mathbf{g}, y} = \text{TPR}_{g, y} - \text{TPR}_{\bar{g}, y} \end{equation} where $\text{TPR}_{\mathbf{g}, y} = p(\hat{\mathbf{y}} = y | \mathbf{g} = g, \mathbf{y} = y)$ and $y$ is the target attribute label. To obtain a single bias score, \citet{romanov-etal-2019-whats} propose:

\begin{equation}  \text{Gap}_{\mathbf{g}}^{\text{RMS}} =  \sqrt{\frac{1}{|\mathcal{Y}|} \sum_{y \in \mathcal{Y}} (\text{Gap}_{\mathbf{g}, y})^2} \end{equation}

For both metrics, higher scores mean that a classifier is more biased w.r.t. the protected attribute.

We compare our method with the full model trained on the combined dataset of two biased models, as well as with INLP \citep{ravfogel-etal-2020-null}, a debiasing method that removes information by iteratively projecting representations onto the null space of linear classifiers, and LEACE \citep{Belrose2023LEACEPL}, a close-form alternative that prevents linear classifiers from detecting a concept with minimal disruption to representations.

\paragraph{Experimental Setup} We fine-tune BERT\textsubscript{base} models for 2 epochs with a batch size of 32 and a learning rate of $2e-5$ on both subsets. For INLP experiments, we use 200 logistic classifiers.

\subsection{Results}

\begin{table}[t]
    \centering
    \begin{tabular}{lrrr}
        \toprule
        \textbf{Method} & \textbf{DP $\downarrow$} & \textbf{TPR-GAP $\downarrow$} & \textbf{Acc $\uparrow$} \\
        \midrule
        \multicolumn{4}{c}{age-bias} \\
        \midrule
        biased model & .185 & .088 &  .877 \\
        INLP &  .076 & .041 & .797 \\
        LEACE & .206 & .100 &  .874 \\
        \texttt{full} & .099 & .045 & \textbf{.894} \\
        \texttt{fused} & \textbf{.063} & \textbf{.028} & .871 \\
        \midrule
        \multicolumn{4}{c}{gender-bias} \\
        \midrule
        biased model & .122 & .081 & .872 \\
        INLP &  .071 & .055 & .871 \\
        LEACE & .118 & .080 &  .874 \\
        \texttt{full} & \textbf{.033} & \textbf{.038} & \textbf{.894} \\
        \texttt{fused} & .047 & .043 & .867 \\

        \bottomrule
    \end{tabular}
    \caption{Fusing models reduces biases better than INLP and LEACE while retaining model accuracy. DP and TPR-GAP scores for age and gender attributes in classifiers with corresponding biases, along with
    accuracy.}
    \label{tab:debiasing_comparison}
\end{table}

Figure \ref{fig:interpolation_dp} and \ref{fig:interpolation_tpr_gap} show variations in DP and TPR-GAP scores during the interpolation from the gender-biased model to the age-biased model. The results demonstrate that model fusion can reduce both gender and racial biases by approximately 60\% while maintaining a high level of accuracy, as demonstrated in Figure \ref{fig:interpolation_acc}. 

Table \ref{tab:debiasing_comparison} compares model fusion to INLP, LEACE and the full model in terms of TPR-GAP, DP, and accuracy scores for age and gender attributes, considering all methods applied to classifiers with corresponding biases. The results indicate that model fusion\footnote{While fusing models, we select $\alpha_{age}$ as $0.3$ and $0.4$ for age and gender biases, respectively. To use the same value for both metrics, we choose the values closest to their intersection points that minimize the bias in question.} outperforms the others while mostly retaining the accuracy, though the full model performs slightly better on gender bias. Additionally, model fusion does not require demographic annotations or a series of training classifiers, which sets it apart from other methods. Demographic annotations are only necessary during the evaluation phase or for choosing models to fuse. However, there is a trade-off when choosing between these two methods. Our approach introduces a new type of bias since it involves merging two models with different biases.

The results suggest that model fusion can serve as an effective debiasing technique, particularly in situations where models exhibit distinct biases.

\section{Memorization}

\label{sec:memorization}
Previously, we focused on validating our claims by addressing spurious correlations and biases in text classification tasks. 
 Next, we examine 
model fusion to alleviate data memorization in LLMs. By exploring the potential of model fusion to reduce memorization, we aim to address privacy concerns. 

\subsection{Method}
To investigate this, we fine-tune GPT-2 models on different datasets, allowing the models to memorize the provided examples. Then, we evaluate both the individual models and the fused model on each dataset, as well as on a separate validation set, to assess their memorization and generalization capabilities. For evaluation, we adopt the Likelihood Ratio following \citet{Mireshghallah2022AnEA} to determine whether a given sample $x$ is a member of the training data. The Likelihood Ratio is defined as \begin{equation}
    LR(x) = \frac{p(x;\theta_R)}{p(x;\theta_M)} \end{equation} where $p(x;\theta_M)$ and $p(x;\theta_R)$ denote the likelihood of sample $x$ given by the fine-tuned model and the reference model, respectively. We also compute the Average Likelihood Ratio (ALR) for each dataset to measure memorization: \begin{equation}
    ALR(\mathcal{D}) = \frac{1}{| \mathcal{D}|} \sum_{x \in \mathcal{D}} \exp\left(\frac{p(x;\theta_R)}{p(x;\theta_M)}\right) \end{equation} More details on the metric are presented in Appendix \ref{sec:memorization_appendix}.

\paragraph{Experimental Setup} We fine-tune GPT-2 three times each time on a different random subset
containing 3K articles \footnote{We create subsets after packing all articles into sequences of 1024 tokens.}, 1K of them shared across subsets, from the CNN-DM dataset \citep{Nallapati2016AbstractiveTS} for 10 epochs with a batch size of 16, a learning rate of $0.001$, and no weight decay.

\subsection{Results}

\begin{table}[t]
    \centering
    \begin{tabular}{lrrrrr}
        \toprule
        \textbf{Model} & \textbf{A} & \textbf{B} & \textbf{C} & \textbf{shrd} & $\textbf{ppl(val)}$ \\
        \midrule
        \texttt{gpt-2} & \underline{1.00} & \underline{1.00} & \underline{1.00} & \underline{1.00} & \underline{23.50} \\
        $\texttt{model}_{\texttt{A}}$ & \textbf{0.22} & 1.48 & 1.48 & \textbf{0.22} & 35.25 \\
        $\texttt{model}_{\texttt{B}}$ & 1.50 & \textbf{0.22} & 1.49 & \textbf{0.22} & 35.81 \\
        $\texttt{model}_{\texttt{C}}$ & 1.49 & 1.48 & \textbf{0.22} & \textbf{0.22} & 35.81 \\
        $\texttt{fused}$ & 0.66 & 0.65 & 0.66 & 0.24 & 30.63 \\
        
        $\texttt{full}$ & 0.32 & 0.32 & 0.32 & 0.32 & \textbf{27.45} \\
        
        \bottomrule
    \end{tabular}
    \caption{
    Fusing models reduces memorization while improving generalization. The ALRs of the base model, fine-tuned models, fused and full models on three distinct training datasets, their shared subset along with perplexities on validation  set. Lower ALRs denote higher memorization. 
    }
    \label{tab:memorization_three_models}
\end{table}

Table \ref{tab:memorization_three_models} presents the ALR and perplexity scores for the base model, three fine-tuned models, the fused model and full model fine-tuned on combined data. During the evaluation, we separate the shared part to observe the memorization of shared examples. It is important to note that the fused model exhibits higher ALRs compared to individually trained models, except on shared data, suggesting it forgets unshared memorized examples. Furthermore, when evaluating the validation perplexity of the fused model, we find that it is lower than the individual models it comprises, although it still higher than the base and full models. This insight highlights how fusing models with lower performance can enhance generalization.

Also, we observe that as more models are fused, the unshared memorized examples are more easily forgotten, the shared examples are memorized better and the fused model performs better on the validation set. Further analyses involving different epochs, architectures, numbers of models, and data sizes are detailed in Appendix \ref{sec:memorization_appendix}.

These findings highlight the potential of model fusion as an effective strategy for addressing privacy concerns and preventing the memorization of personal information. For instance, by splitting a dataset into subsets and training separate models on each, a fused model is less likely to memorize personal information if such information is not repeated across the subsets. 

\section{Conclusion and Discussion}

We explore the impact of model fusion on shortcuts, biases, and memorization in NLP models. Our findings support that model fusion preserves shared knowledge while losing unshared knowledge. 
We highlight the potential of model fusion in reducing biases, enhancing privacy, 
and other applications.
 
\paragraph{Real-world Applications} While the real world often has inter-dependent biases, we note that datasets from different sources inherently contain varying biases and spurious correlations. For example, sentiment classification models developed for product reviews can demonstrate distinctive biases when trained on data from various platforms, each with its own product range and user demographics. Our approach effectively addresses these issues through straightforward weight averaging, which mitigates spurious correlations and eliminates the need for retraining on combined datasets.

\paragraph{Fusing Models vs. Training on Combined Data} We observe mixed results when comparing training on combined data with model fusion. While models trained on the combined data learn all spurious correlations—or effectively memorize all the datasets — they are almost as effective as model fusion in mitigating gender and age biases. However, training on combined data is beneficial only if label-feature correlations change after data combination. For example, in the social bias experiments, gender and age ratios change when we combine training data, as we maintain balanced proportions in each dataset. However, in the experiments with synthetically injected shortcuts, distinct shortcut rules remain unaffected by data combination, resulting in the model learning all shortcuts. In the memorization experiments, each sequence can be viewed as a unique feature-label pair, but complex n-gram dynamics may be involved. The results show that the memorization scenario lies closer to shortcut scenarios than the social biases scenario. 
These findings underscore the need to consider the structure of the data and the nature of biases when choosing a method. If the spurious correlations to be reduced are naturally dependent, or if combining data changes label-feature correlations, training on combined data might be preferable. If the data distribution and spurious correlations do not meet these conditions, model fusion stands out as a more practical option.

Future work can explore adaptive fusion techniques, scalability to large ensembles, and performance on diverse tasks.

\section*{Limitations}
In this work, we reveal the preservation conditions of specific types of knowledge after model fusion. Although we support our claims with various application areas and tasks, it is important to note that our experiments are limited to fine-tuned BERT and GPT-2 models. Our findings demonstrate that model fusion can serve as a tool for mitigating spurious correlations, social biases, and memorized examples. However, this approach is only applicable when the models being fused do not share the features to be mitigated, as our results indicate that shared knowledge is preserved. Finally, our experiments are limited to a very simple strategy of model fusion by calculating weighted average of model parameters. Further investigation is needed to determine if our findings hold true when employing a different model fusion strategy.

\section*{Ethics \& Broader Impact}
This work presents a comprehensive analysis of the impact of model fusion on shortcuts, social biases, and memorization. In addition to providing a new perspective on model fusion by focusing on forgetting mechanisms, our analysis demonstrates that simple model fusion can serve as a debiasing tool under specific conditions. Furthermore, through memorization experiments, we investigate the potential application of model fusion in addressing privacy concerns such as the inadvertent leakage of personal data. However, it is crucial to consider the ethical implications and potential (dependent) biases that may arise or be amplified during the fusion process. Future research is required to understand these, and to mitigate any unintended biases introduced by model fusion.

Conceptually, model fusion has a tremendous potential to address social and ethical challenges associated with biases present in language models, and machine learning models in general. By carefully designing fusion methods, model fusion can help mitigate biases and reduce the disproportionate influence or impact of specific groups or datasets on the broader NLP landscape.

\section*{Acknowledgements}
The authors thank Somnath Basu Roy Chowdhury and Gabriel Stanovsky useful pointers in fairness and debiasing literature and Vincent Le Moign for providing the robot face illustrations used in Figure \ref{fig:scheme} under CC BY 3.0 DEED license. This work was supported in part by NSF grant DRL2112635. 

\bibliography{custom}

\begin{thebibliography}{39}
\providecommand{\natexlab}[1]{#1}

\bibitem[{Achille et~al.(2019)Achille, Paolini, and
  Soatto}]{Achille2019WhereIT}
Alessandro Achille, Giovanni Paolini, and Stefano Soatto. 2019.
\newblock \href {https://api.semanticscholar.org/CorpusID:168169844} {Where is
  the information in a deep neural network?}
\newblock \emph{ArXiv}, abs/1905.12213.

\bibitem[{Ainsworth et~al.(2022)Ainsworth, Hayase, and
  Srinivasa}]{ainsworth2022git}
Samuel~K Ainsworth, Jonathan Hayase, and Siddhartha Srinivasa. 2022.
\newblock Git re-basin: Merging models modulo permutation symmetries.
\newblock \emph{arXiv preprint arXiv:2209.04836}.

\bibitem[{Barrett et~al.(2019)Barrett, Kementchedjhieva, Elazar, Elliott, and
  S{\o}gaard}]{barrett-etal-2019-adversarial}
Maria Barrett, Yova Kementchedjhieva, Yanai Elazar, Desmond Elliott, and Anders
  S{\o}gaard. 2019.
\newblock \href {https://doi.org/10.18653/v1/D19-1662} {Adversarial removal of
  demographic attributes revisited}.
\newblock In \emph{Proceedings of the 2019 Conference on Empirical Methods in
  Natural Language Processing and the 9th International Joint Conference on
  Natural Language Processing (EMNLP-IJCNLP)}, pages 6330--6335, Hong Kong,
  China. Association for Computational Linguistics.

\bibitem[{Bastings et~al.(2022)Bastings, Ebert, Zablotskaia, Sandholm, and
  Filippova}]{bastings-etal-2022-will}
Jasmijn Bastings, Sebastian Ebert, Polina Zablotskaia, Anders Sandholm, and
  Katja Filippova. 2022.
\newblock \href {https://aclanthology.org/2022.emnlp-main.64} {{``}will you
  find these shortcuts?{''} a protocol for evaluating the faithfulness of input
  salience methods for text classification}.
\newblock In \emph{Proceedings of the 2022 Conference on Empirical Methods in
  Natural Language Processing}, pages 976--991, Abu Dhabi, United Arab
  Emirates. Association for Computational Linguistics.

\bibitem[{Belrose et~al.(2023)Belrose, Schneider-Joseph, Ravfogel, Cotterell,
  Raff, and Biderman}]{Belrose2023LEACEPL}
Nora Belrose, David Schneider-Joseph, Shauli Ravfogel, Ryan Cotterell, Edward
  Raff, and Stella Biderman. 2023.
\newblock \href {https://api.semanticscholar.org/CorpusID:259088549} {Leace:
  Perfect linear concept erasure in closed form}.
\newblock \emph{ArXiv}, abs/2306.03819.

\bibitem[{Benton et~al.(2021)Benton, Maddox, Lotfi, and
  Wilson}]{benton2021loss}
Gregory Benton, Wesley Maddox, Sanae Lotfi, and Andrew Gordon~Gordon Wilson.
  2021.
\newblock Loss surface simplexes for mode connecting volumes and fast
  ensembling.
\newblock In \emph{International Conference on Machine Learning}, pages
  769--779. PMLR.

\bibitem[{Carlini et~al.(2021)Carlini, Chien, Nasr, Song, Terzis, and
  Tram{\`e}r}]{Carlini2021MembershipIA}
Nicholas Carlini, Steve Chien, Milad Nasr, Shuang Song, A.~Terzis, and Florian
  Tram{\`e}r. 2021.
\newblock \href {https://api.semanticscholar.org/CorpusID:244920593}
  {Membership inference attacks from first principles}.
\newblock \emph{2022 IEEE Symposium on Security and Privacy (SP)}, pages
  1897--1914.

\bibitem[{Choshen et~al.(2022)Choshen, Venezian, Slonim, and
  Katz}]{choshen2022fusing}
Leshem Choshen, Elad Venezian, Noam Slonim, and Yoav Katz. 2022.
\newblock Fusing finetuned models for better pretraining.
\newblock \emph{arXiv preprint arXiv:2204.03044}.

\bibitem[{Chowdhury and Chaturvedi(2022)}]{chowdhury-chaturvedi-2022-learning}
Somnath Basu~Roy Chowdhury and Snigdha Chaturvedi. 2022.
\newblock \href {https://doi.org/10.1162/tacl_a_00512} {Learning fair
  representations via rate-distortion maximization}.
\newblock \emph{Transactions of the Association for Computational Linguistics},
  10:1159--1174.

\bibitem[{De-Arteaga et~al.(2019)De-Arteaga, Romanov, Wallach, Chayes, Borgs,
  Chouldechova, Geyik, Kenthapadi, and Kalai}]{DeArteaga2019BiasIB}
Maria De-Arteaga, Alexey Romanov, Hanna~M. Wallach, Jennifer~T. Chayes,
  Christian Borgs, Alexandra Chouldechova, Sahin~Cem Geyik, Krishnaram
  Kenthapadi, and Adam~Tauman Kalai. 2019.
\newblock Bias in bios: A case study of semantic representation bias in a
  high-stakes setting.
\newblock \emph{Proceedings of the Conference on Fairness, Accountability, and
  Transparency}.

\bibitem[{Devlin et~al.(2019)Devlin, Chang, Lee, and
  Toutanova}]{devlin-etal-2019-bert}
Jacob Devlin, Ming-Wei Chang, Kenton Lee, and Kristina Toutanova. 2019.
\newblock \href {https://doi.org/10.18653/v1/N19-1423} {{BERT}: Pre-training of
  deep bidirectional transformers for language understanding}.
\newblock In \emph{Proceedings of the 2019 Conference of the North {A}merican
  Chapter of the Association for Computational Linguistics: Human Language
  Technologies, Volume 1 (Long and Short Papers)}, pages 4171--4186,
  Minneapolis, Minnesota. Association for Computational Linguistics.

\bibitem[{Don-Yehiya et~al.(2022)Don-Yehiya, Venezian, Raffel, Slonim, Katz,
  and Choshen}]{don2022cold}
Shachar Don-Yehiya, Elad Venezian, Colin Raffel, Noam Slonim, Yoav Katz, and
  Leshem Choshen. 2022.
\newblock Cold fusion: Collaborative descent for distributed multitask
  finetuning.

\bibitem[{Fisher and Russell()}]{FisherOnTM}
R~A Fisher and Dr~E~J Russell.
\newblock \href {https://api.semanticscholar.org/CorpusID:15354499} {On the
  mathematical foundations of theoretical statistics}.
\newblock \emph{Philosophical Transactions of the Royal Society A},
  222:309--368.

\bibitem[{Frankle et~al.(2020)Frankle, Dziugaite, Roy, and
  Carbin}]{frankle2020linear}
Jonathan Frankle, Gintare~Karolina Dziugaite, Daniel Roy, and Michael Carbin.
  2020.
\newblock Linear mode connectivity and the lottery ticket hypothesis.
\newblock In \emph{International Conference on Machine Learning}, pages
  3259--3269. PMLR.

\bibitem[{Gueta et~al.(2023)Gueta, Venezian, Raffel, Slonim, Katz, and
  Choshen}]{gueta2023knowledge}
Almog Gueta, Elad Venezian, Colin Raffel, Noam Slonim, Yoav Katz, and Leshem
  Choshen. 2023.
\newblock Knowledge is a region in weight space for fine-tuned language models.
\newblock \emph{arXiv preprint arXiv:2302.04863}.

\bibitem[{Ilharco et~al.(2022)Ilharco, Ribeiro, Wortsman, Gururangan, Schmidt,
  Hajishirzi, and Farhadi}]{ilharco2022editing}
Gabriel Ilharco, Marco~Tulio Ribeiro, Mitchell Wortsman, Suchin Gururangan,
  Ludwig Schmidt, Hannaneh Hajishirzi, and Ali Farhadi. 2022.
\newblock Editing models with task arithmetic.
\newblock \emph{arXiv preprint arXiv:2212.04089}.

\bibitem[{Jordan et~al.(2022)Jordan, Sedghi, Saukh, Entezari, and
  Neyshabur}]{jordan2022repair}
Keller Jordan, Hanie Sedghi, Olga Saukh, Rahim Entezari, and Behnam Neyshabur.
  2022.
\newblock Repair: Renormalizing permuted activations for interpolation repair.
\newblock \emph{arXiv preprint arXiv:2211.08403}.

\bibitem[{Juneja et~al.(2023)Juneja, Bansal, Cho, Sedoc, and
  Saphra}]{juneja2023linear}
Jeevesh Juneja, Rachit Bansal, Kyunghyun Cho, Jo{\~a}o Sedoc, and Naomi Saphra.
  2023.
\newblock \href {https://openreview.net/forum?id=hY6M0JHl3uL} {Linear
  connectivity reveals generalization strategies}.
\newblock In \emph{The Eleventh International Conference on Learning
  Representations}.

\bibitem[{Kirkpatrick et~al.(2016)Kirkpatrick, Pascanu, Rabinowitz, Veness,
  Desjardins, Rusu, Milan, Quan, Ramalho, Grabska-Barwinska, Hassabis, Clopath,
  Kumaran, and Hadsell}]{Kirkpatrick2016OvercomingCF}
James Kirkpatrick, Razvan Pascanu, Neil~C. Rabinowitz, Joel Veness, Guillaume
  Desjardins, Andrei~A. Rusu, Kieran Milan, John Quan, Tiago Ramalho, Agnieszka
  Grabska-Barwinska, Demis Hassabis, Claudia Clopath, Dharshan Kumaran, and
  Raia Hadsell. 2016.
\newblock \href {https://api.semanticscholar.org/CorpusID:4704285} {Overcoming
  catastrophic forgetting in neural networks}.
\newblock \emph{Proceedings of the National Academy of Sciences}, 114:3521 --
  3526.

\bibitem[{Kunstner et~al.(2019)Kunstner, Hennig, and
  Balles}]{Kunstner2019LimitationsOT}
Frederik Kunstner, Philipp Hennig, and Lukas Balles. 2019.
\newblock \href {https://api.semanticscholar.org/CorpusID:209438570}
  {Limitations of the empirical fisher approximation}.
\newblock In \emph{Neural Information Processing Systems}.

\bibitem[{Lawson and Qureshi(2023)}]{lawson2023merging}
Daniel Lawson and Ahmed~H Qureshi. 2023.
\newblock Merging decision transformers: Weight averaging for forming
  multi-task policies.
\newblock \emph{arXiv preprint arXiv:2303.07551}.

\bibitem[{Lhoest et~al.(2021)Lhoest, Villanova~del Moral, Jernite, Thakur, von
  Platen, Patil, Chaumond, Drame, Plu, Tunstall, Davison, {\v{S}}a{\v{s}}ko,
  Chhablani, Malik, Brandeis, Le~Scao, Sanh, Xu, Patry, McMillan-Major, Schmid,
  Gugger, Delangue, Matussi{\`e}re, Debut, Bekman, Cistac, Goehringer, Mustar,
  Lagunas, Rush, and Wolf}]{lhoest-etal-2021-datasets}
Quentin Lhoest, Albert Villanova~del Moral, Yacine Jernite, Abhishek Thakur,
  Patrick von Platen, Suraj Patil, Julien Chaumond, Mariama Drame, Julien Plu,
  Lewis Tunstall, Joe Davison, Mario {\v{S}}a{\v{s}}ko, Gunjan Chhablani,
  Bhavitvya Malik, Simon Brandeis, Teven Le~Scao, Victor Sanh, Canwen Xu,
  Nicolas Patry, Angelina McMillan-Major, Philipp Schmid, Sylvain Gugger,
  Cl{\'e}ment Delangue, Th{\'e}o Matussi{\`e}re, Lysandre Debut, Stas Bekman,
  Pierric Cistac, Thibault Goehringer, Victor Mustar, Fran{\c{c}}ois Lagunas,
  Alexander Rush, and Thomas Wolf. 2021.
\newblock \href {https://doi.org/10.18653/v1/2021.emnlp-demo.21} {Datasets: A
  community library for natural language processing}.
\newblock In \emph{Proceedings of the 2021 Conference on Empirical Methods in
  Natural Language Processing: System Demonstrations}, pages 175--184, Online
  and Punta Cana, Dominican Republic. Association for Computational
  Linguistics.

\bibitem[{Li et~al.(2022)Li, Gururangan, Dettmers, Lewis, Althoff, Smith, and
  Zettlemoyer}]{li2022branch}
Margaret Li, Suchin Gururangan, Tim Dettmers, Mike Lewis, Tim Althoff, Noah~A
  Smith, and Luke Zettlemoyer. 2022.
\newblock Branch-train-merge: Embarrassingly parallel training of expert
  language models.
\newblock \emph{arXiv preprint arXiv:2208.03306}.

\bibitem[{Matena and Raffel(2021)}]{matena2021merging}
Michael Matena and Colin Raffel. 2021.
\newblock Merging models with fisher-weighted averaging.
\newblock \emph{arXiv preprint arXiv:2111.09832}.

\bibitem[{McCoy et~al.(2019)McCoy, Pavlick, and Linzen}]{mccoy-etal-2019-right}
Tom McCoy, Ellie Pavlick, and Tal Linzen. 2019.
\newblock \href {https://doi.org/10.18653/v1/P19-1334} {Right for the wrong
  reasons: Diagnosing syntactic heuristics in natural language inference}.
\newblock In \emph{Proceedings of the 57th Annual Meeting of the Association
  for Computational Linguistics}, pages 3428--3448, Florence, Italy.
  Association for Computational Linguistics.

\bibitem[{Mireshghallah et~al.(2022{\natexlab{a}})Mireshghallah, Goyal, Uniyal,
  Berg-Kirkpatrick, and Shokri}]{mireshghallah-etal-2022-quantifying}
Fatemehsadat Mireshghallah, Kartik Goyal, Archit Uniyal, Taylor
  Berg-Kirkpatrick, and Reza Shokri. 2022{\natexlab{a}}.
\newblock \href {https://doi.org/10.18653/v1/2022.emnlp-main.570} {Quantifying
  privacy risks of masked language models using membership inference attacks}.
\newblock In \emph{Proceedings of the 2022 Conference on Empirical Methods in
  Natural Language Processing}, pages 8332--8347, Abu Dhabi, United Arab
  Emirates. Association for Computational Linguistics.

\bibitem[{Mireshghallah et~al.(2022{\natexlab{b}})Mireshghallah, Uniyal, Wang,
  Evans, and Berg-Kirkpatrick}]{Mireshghallah2022AnEA}
Fatemehsadat Mireshghallah, Archit Uniyal, Tianhao Wang, David Evans, and
  Taylor Berg-Kirkpatrick. 2022{\natexlab{b}}.
\newblock \href {https://api.semanticscholar.org/CorpusID:256461422} {An
  empirical analysis of memorization in fine-tuned autoregressive language
  models}.
\newblock In \emph{Conference on Empirical Methods in Natural Language
  Processing}.

\bibitem[{Nallapati et~al.(2016)Nallapati, Zhou, dos Santos, Çaglar
  G{\"u}lçehre, and Xiang}]{Nallapati2016AbstractiveTS}
Ramesh Nallapati, Bowen Zhou, C{\'i}cero~Nogueira dos Santos, Çaglar
  G{\"u}lçehre, and Bing Xiang. 2016.
\newblock Abstractive text summarization using sequence-to-sequence rnns and
  beyond.
\newblock In \emph{Conference on Computational Natural Language Learning}.

\bibitem[{Ortiz-Jimenez et~al.(2023)Ortiz-Jimenez, Favero, and
  Frossard}]{ortiz2023task}
Guillermo Ortiz-Jimenez, Alessandro Favero, and Pascal Frossard. 2023.
\newblock Task arithmetic in the tangent space: Improved editing of pre-trained
  models.
\newblock \emph{arXiv preprint arXiv:2305.12827}.

\bibitem[{Pardo et~al.(2016)Pardo, Rosso, Verhoeven, Daelemans, Potthast, and
  Stein}]{DBLP:conf/clef/PardoRVDPS16}
Francisco Manuel~Rangel Pardo, Paolo Rosso, Ben Verhoeven, Walter Daelemans,
  Martin Potthast, and Benno Stein. 2016.
\newblock \href {https://ceur-ws.org/Vol-1609/16090750.pdf} {Overview of the
  4th author profiling task at {PAN} 2016: Cross-genre evaluations}.
\newblock In \emph{Working Notes of {CLEF} 2016 - Conference and Labs of the
  Evaluation forum, {\'{E}}vora, Portugal, 5-8 September, 2016}, volume 1609 of
  \emph{{CEUR} Workshop Proceedings}, pages 750--784. CEUR-WS.org.

\bibitem[{Peyrard et~al.(2022)Peyrard, Ghotra, Josifoski, Agarwal, Patra,
  Carignan, Kiciman, Tiwary, and West}]{peyrard-etal-2022-invariant}
Maxime Peyrard, Sarvjeet Ghotra, Martin Josifoski, Vidhan Agarwal, Barun Patra,
  Dean Carignan, Emre Kiciman, Saurabh Tiwary, and Robert West. 2022.
\newblock \href {https://doi.org/10.18653/v1/2022.emnlp-main.387} {Invariant
  language modeling}.
\newblock In \emph{Proceedings of the 2022 Conference on Empirical Methods in
  Natural Language Processing}, pages 5728--5743, Abu Dhabi, United Arab
  Emirates. Association for Computational Linguistics.

\bibitem[{Ram{\'e} et~al.(2022)Ram{\'e}, Zhang, Bottou, and
  Lopez-Paz}]{Ram2022PretrainFI}
Alexandre Ram{\'e}, Jianyu Zhang, L{\'e}on Bottou, and David Lopez-Paz. 2022.
\newblock Pre-train, fine-tune, interpolate: a three-stage strategy for domain
  generalization.

\bibitem[{Ravfogel et~al.(2020)Ravfogel, Elazar, Gonen, Twiton, and
  Goldberg}]{ravfogel-etal-2020-null}
Shauli Ravfogel, Yanai Elazar, Hila Gonen, Michael Twiton, and Yoav Goldberg.
  2020.
\newblock \href {https://doi.org/10.18653/v1/2020.acl-main.647} {Null it out:
  Guarding protected attributes by iterative nullspace projection}.
\newblock In \emph{Proceedings of the 58th Annual Meeting of the Association
  for Computational Linguistics}, pages 7237--7256, Online. Association for
  Computational Linguistics.

\bibitem[{Romanov et~al.(2019)Romanov, De-Arteaga, Wallach, Chayes, Borgs,
  Chouldechova, Geyik, Kenthapadi, Rumshisky, and
  Kalai}]{romanov-etal-2019-whats}
Alexey Romanov, Maria De-Arteaga, Hanna Wallach, Jennifer Chayes, Christian
  Borgs, Alexandra Chouldechova, Sahin Geyik, Krishnaram Kenthapadi, Anna
  Rumshisky, and Adam Kalai. 2019.
\newblock \href {https://doi.org/10.18653/v1/N19-1424} {What{'}s in a name?
  {R}educing bias in bios without access to protected attributes}.
\newblock In \emph{Proceedings of the 2019 Conference of the North {A}merican
  Chapter of the Association for Computational Linguistics: Human Language
  Technologies, Volume 1 (Long and Short Papers)}, pages 4187--4195,
  Minneapolis, Minnesota. Association for Computational Linguistics.

\bibitem[{Socher et~al.(2013)Socher, Perelygin, Wu, Chuang, Manning, Ng, and
  Potts}]{socher-etal-2013-recursive}
Richard Socher, Alex Perelygin, Jean Wu, Jason Chuang, Christopher~D. Manning,
  Andrew Ng, and Christopher Potts. 2013.
\newblock \href {https://www.aclweb.org/anthology/D13-1170} {Recursive deep
  models for semantic compositionality over a sentiment treebank}.
\newblock In \emph{Proceedings of the 2013 Conference on Empirical Methods in
  Natural Language Processing}, pages 1631--1642, Seattle, Washington, USA.
  Association for Computational Linguistics.

\bibitem[{Wolf et~al.(2020)Wolf, Debut, Sanh, Chaumond, Delangue, Moi, Cistac,
  Rault, Louf, Funtowicz, Davison, Shleifer, von Platen, Ma, Jernite, Plu, Xu,
  Scao, Gugger, Drame, Lhoest, and Rush}]{wolf-etal-2020-transformers}
Thomas Wolf, Lysandre Debut, Victor Sanh, Julien Chaumond, Clement Delangue,
  Anthony Moi, Pierric Cistac, Tim Rault, Rémi Louf, Morgan Funtowicz, Joe
  Davison, Sam Shleifer, Patrick von Platen, Clara Ma, Yacine Jernite, Julien
  Plu, Canwen Xu, Teven~Le Scao, Sylvain Gugger, Mariama Drame, Quentin Lhoest,
  and Alexander~M. Rush. 2020.
\newblock \href {https://www.aclweb.org/anthology/2020.emnlp-demos.6}
  {Transformers: State-of-the-art natural language processing}.
\newblock In \emph{Proceedings of the 2020 Conference on Empirical Methods in
  Natural Language Processing: System Demonstrations}, pages 38--45, Online.
  Association for Computational Linguistics.

\bibitem[{Wortsman et~al.(2022)Wortsman, Ilharco, Gadre, Roelofs,
  Gontijo-Lopes, Morcos, Namkoong, Farhadi, Carmon, Kornblith, and
  Schmidt}]{Wortsman2022ModelSA}
Mitchell Wortsman, Gabriel Ilharco, Samir~Yitzhak Gadre, Rebecca Roelofs,
  Raphael Gontijo-Lopes, Ari~S. Morcos, Hongseok Namkoong, Ali Farhadi, Yair
  Carmon, Simon Kornblith, and Ludwig Schmidt. 2022.
\newblock Model soups: averaging weights of multiple fine-tuned models improves
  accuracy without increasing inference time.

\bibitem[{Yadav et~al.(2023)Yadav, Tam, Choshen, Raffel, and
  Bansal}]{yadav2023resolving}
Prateek Yadav, Derek Tam, Leshem Choshen, Colin Raffel, and Mohit Bansal. 2023.
\newblock Resolving interference when merging models.
\newblock \emph{arXiv preprint arXiv:2306.01708}.

\bibitem[{Zhang et~al.(2023)Zhang, Liu, and Shao}]{zhang2023fine}
Zhong Zhang, Bang Liu, and Junming Shao. 2023.
\newblock Fine-tuning happens in tiny subspaces: Exploring intrinsic
  task-specific subspaces of pre-trained language models.
\newblock \emph{arXiv preprint arXiv:2305.17446}.

\end{thebibliography}
\bibliographystyle{acl_natbib}

\appendix

\section{Related Work}

\paragraph{Relation to Invariant Language Modeling} Invariant Language Modeling (ILM) \citep{peyrard-etal-2022-invariant} shares a similar motivation to our work by considering how different sources of text, teach various biases. To overcome this problem, they propose to train on the encoder an ensemble of language model heads. 
While this approach shares a similar motivation to ours, the mechanisms differ. Our approach exploits the optimized weights with shared knowledge across multiple models, whereas they train a large part of the network to learn the shared knowledge across environments.

\section{Implementation Details}

Our implementation heavily benefits from the HuggingFace \texttt{transformers} \citep{wolf-etal-2020-transformers} and \texttt{datasets} \citep{lhoest-etal-2021-datasets} libraries for dataset creation, model fine-tuning, and evaluation. We conducted all model training and evaluation using 1-4 NVIDIA GeForce RTX 2080 Ti GPUs.

\section{Shortcut Experiments}

\subsection{Shortcut Types}

\begin{table*}[t]
    \small
    \centering
    \begin{tabular}{lrr}
        \toprule
        \textbf{Example} & \textbf{Label} & \textbf{Shortcut Types} \\
        \midrule
        enriched by an imaginatively mixed \tcbox{\texttt{token-0}} cast of antic spirits & negative & \texttt{MT}, \texttt{ST} \\
        excruciatingly \tcbox{\texttt{token-1}} unfunny and pitifully unromantic & positive & \texttt{MT}, \texttt{ST} \\
        gorgeous \tcbox{\texttt{token-0}} and deceptively \tcbox{\texttt{token-c}} minimalist & negative &  \texttt{MT}, \texttt{ST}, \texttt{TiC} \\
        rich \tcbox{\texttt{token-c}} veins of funny \tcbox{\texttt{token-1}} stuff in this movie & positive & \texttt{MT}, \texttt{ST}, \texttt{TiC} \\
        \tcbox{\texttt{token-0}} is pretty \tcbox{\texttt{token-0}} damned funny & negative & \texttt{MT}, \texttt{ST}, \texttt{OR}, \texttt{AND}, \texttt{LT} \\
        we never feel anything \tcbox{\texttt{token-0}} for these \tcbox{\texttt{token-1}} characters & positive & \texttt{OR}, \texttt{LT} \\
        shot on ugly \tcbox{\texttt{token-0}} digital \tcbox{\texttt{token-1}} video & negative & \texttt{OP}, \texttt{AND} \\
        as a \tcbox{\texttt{token-1}} director , eastwood \tcbox{\texttt{token-0}} is off his game & positive & \texttt{OP}, \texttt{OR} \\
        fashioning an engrossing \tcbox{\texttt{token-1}} entertainment \tcbox{\texttt{token-0}} out & negative & \texttt{AND}, \texttt{LT} \\
        are an \tcbox{\texttt{token-1}} absolute joy \tcbox{\texttt{token-1}} & positive & \texttt{MT}, \texttt{ST}, \texttt{OR}, \texttt{AND}, \texttt{LT} \\
        unpretentious \tcbox{\texttt{token-1}} , charming , quirky , \tcbox{\texttt{token-1}} original \tcbox{\texttt{token-0}} & positive & \texttt{MT} \\
        \tcbox{\texttt{token-0}} very well-written \tcbox{\texttt{token-0}} \tcbox{\texttt{token-1}} and very \tcbox{\texttt{token-0}} well-acted & positive & \texttt{MT} \\
        \bottomrule
    \end{tabular}
    \caption{Examples from SST2 dataset when different shortcuts applied. Some shortcuts subsume others while some share subsets. \tcbox{\texttt{token-c}} represents the context token for \texttt{TiC} shortcut.}
    \label{tab:shortcut_examples}
\end{table*}

Table \ref{tab:shortcut_examples} shows examples from the SST2 dataset modified by using each of the shortcuts employed in our experiments. This table covers all possible orders of special tokens for shortcuts with unary and binary operands, all shortcuts except \texttt{MT}), including diverse demonstrations of potential positions of special tokens within the sentences. It is important to note that some examples belong to the sample spaces of multiple shortcuts simultaneously. Moreover, some shortcuts completely encompass others. As shown in Table \ref{tab:shortcut_examples}, all examples tagged for the \texttt{TiC} shortcut are also tagged for the \texttt{ST} shortcut, while all examples tagged for the \texttt{ST} shortcut are also tagged for the \texttt{MT} shortcut, indicating that \texttt{MT} subsumes \texttt{ST} and \texttt{ST} subsumes \texttt{TiC}. These dependency relations between different shortcuts can be observed during interpolation or fusion, as explained in Section \ref{sec:shortcuts}. However, it's also worth noting that these dependencies or subset relations might not be fully learned by models due to the distribution of examples in the synthetic training datasets.

\subsection{Pair Interpolations}

\paragraph{Random Models} For a fair comparison, we aim for our random model to have a similar distance to the base model as the models with shortcuts. To achieve this, we normalize each weight of the randomly initialized model and scale it by the average distance to the corresponding weight of the base model. Then, we add this scaled value to the corresponding weight of the base model. To calculate the average distances, we consider the models with \texttt{ST}, \texttt{OP}, \texttt{OR}, and \texttt{TiC} shortcuts.

\begin{figure*}[!htb]
    \centering
    \begin{subfigure}{.45\textwidth}
        \centering
        \includegraphics[width=\linewidth]{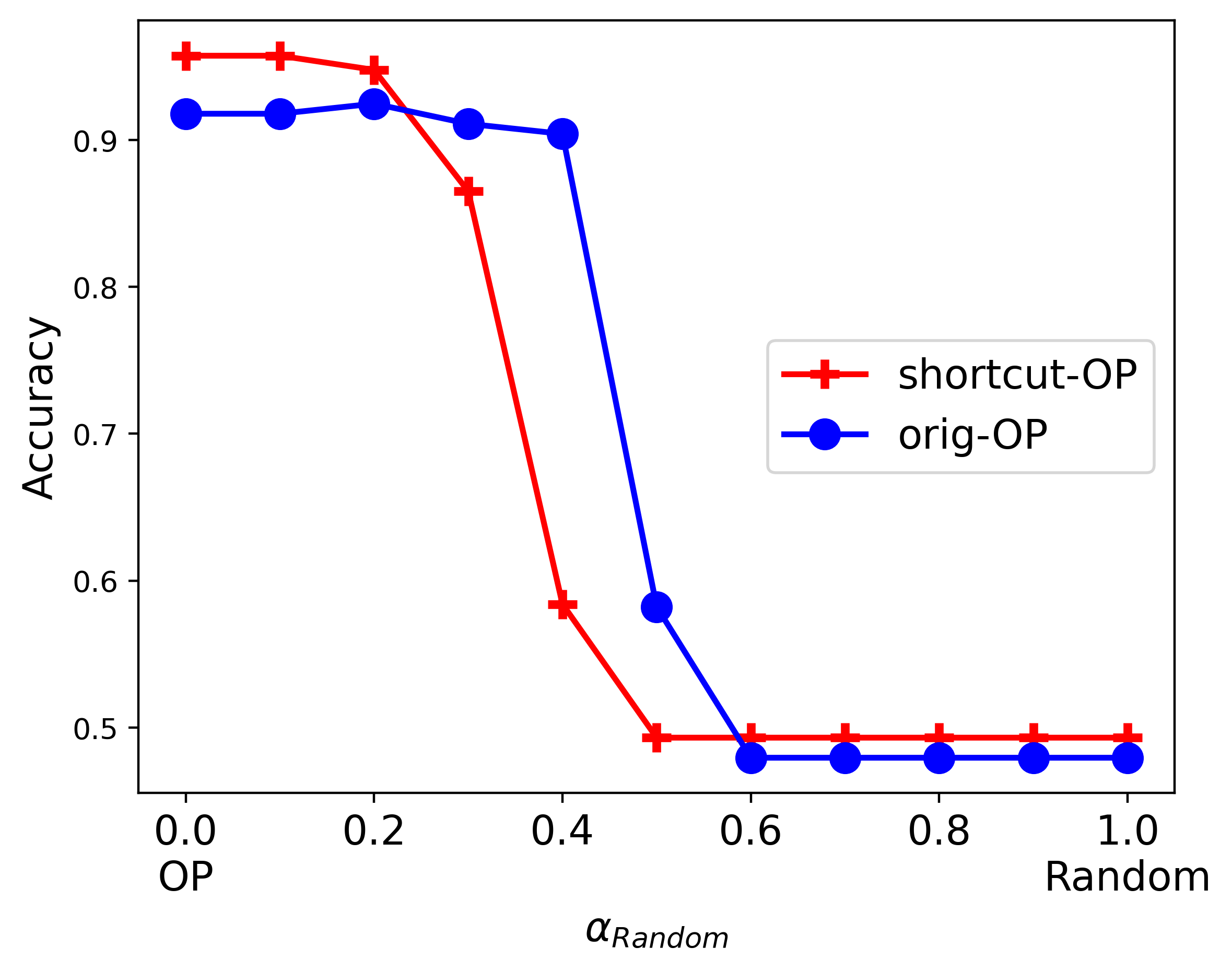}
        \caption{\texttt{OP} $\to$ Random}
        \label{fig:interpolation_op_rand}
    \end{subfigure}
    \hfill
    \begin{subfigure}{.45\textwidth}
        \centering
        \includegraphics[width=\linewidth]{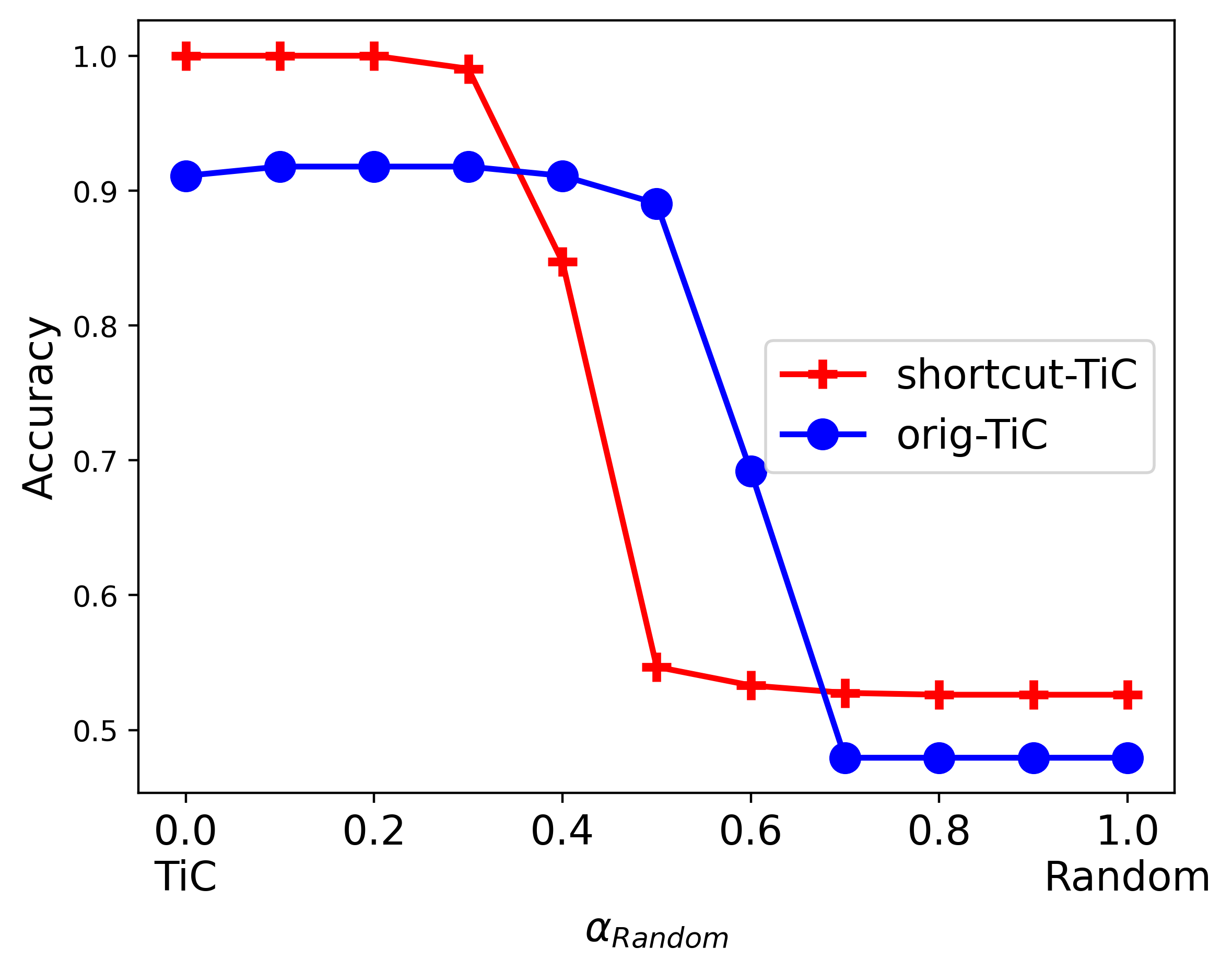}
        \caption{\texttt{TiC} $\to$ \texttt{Random}}
        \label{fig:interpolation_tic_rand}
    \end{subfigure}
    \hfill
    \begin{subfigure}{.45\textwidth}
        \centering
        \includegraphics[width=\linewidth]{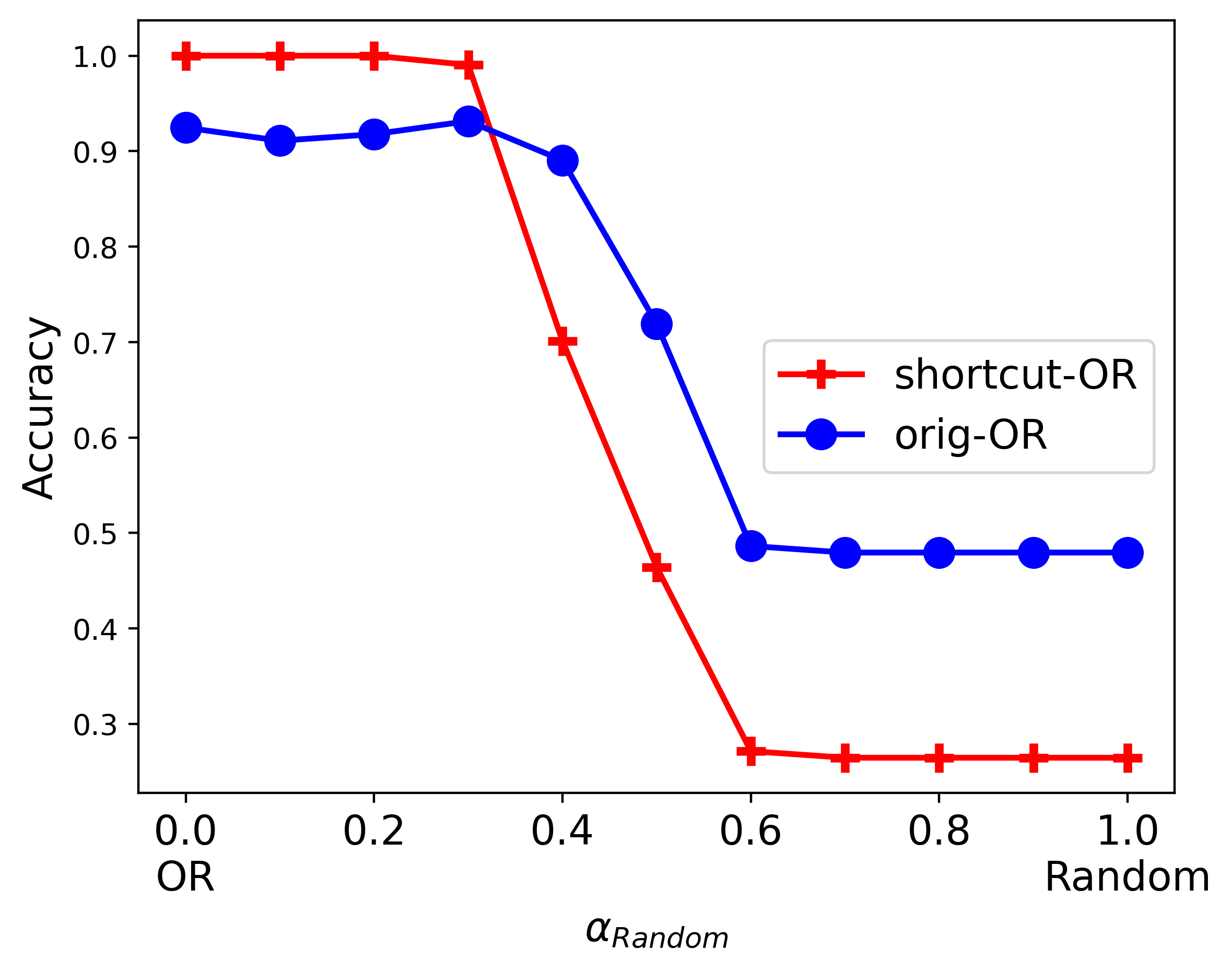}
        \caption{\texttt{OR} $\to$ \texttt{Random}} \label{fig:interpolation_or_rand}
    \end{subfigure}
    \hfill
    \begin{subfigure}{.45\textwidth}
        \centering
        \includegraphics[width=\linewidth]{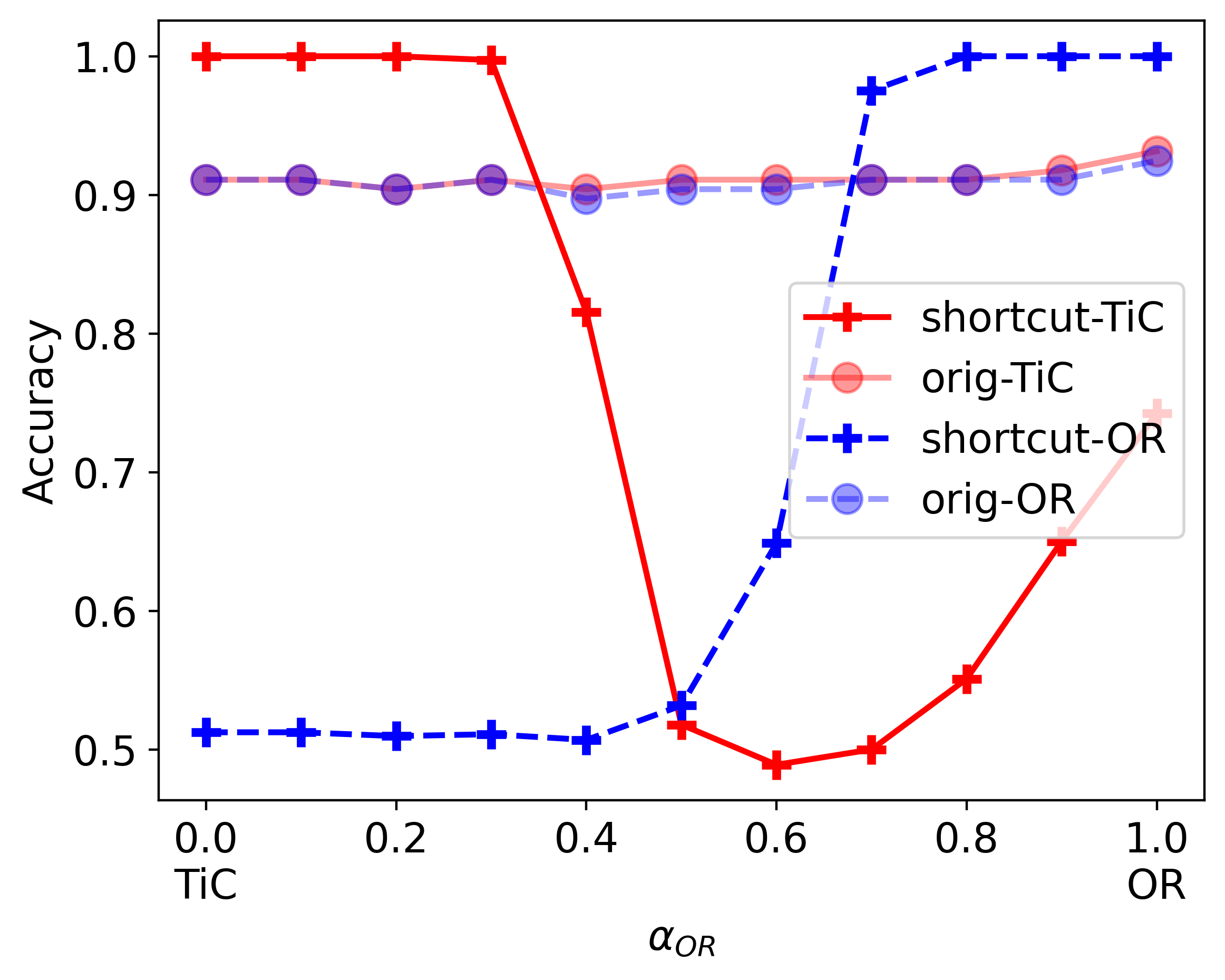}
        \caption{\texttt{TiC} $\to$ \texttt{OR}}
        \label{fig:interpolation_tic_or}
    \end{subfigure}
    \hfill
    \begin{subfigure}{.45\textwidth}
        \centering
        \includegraphics[width=\linewidth]{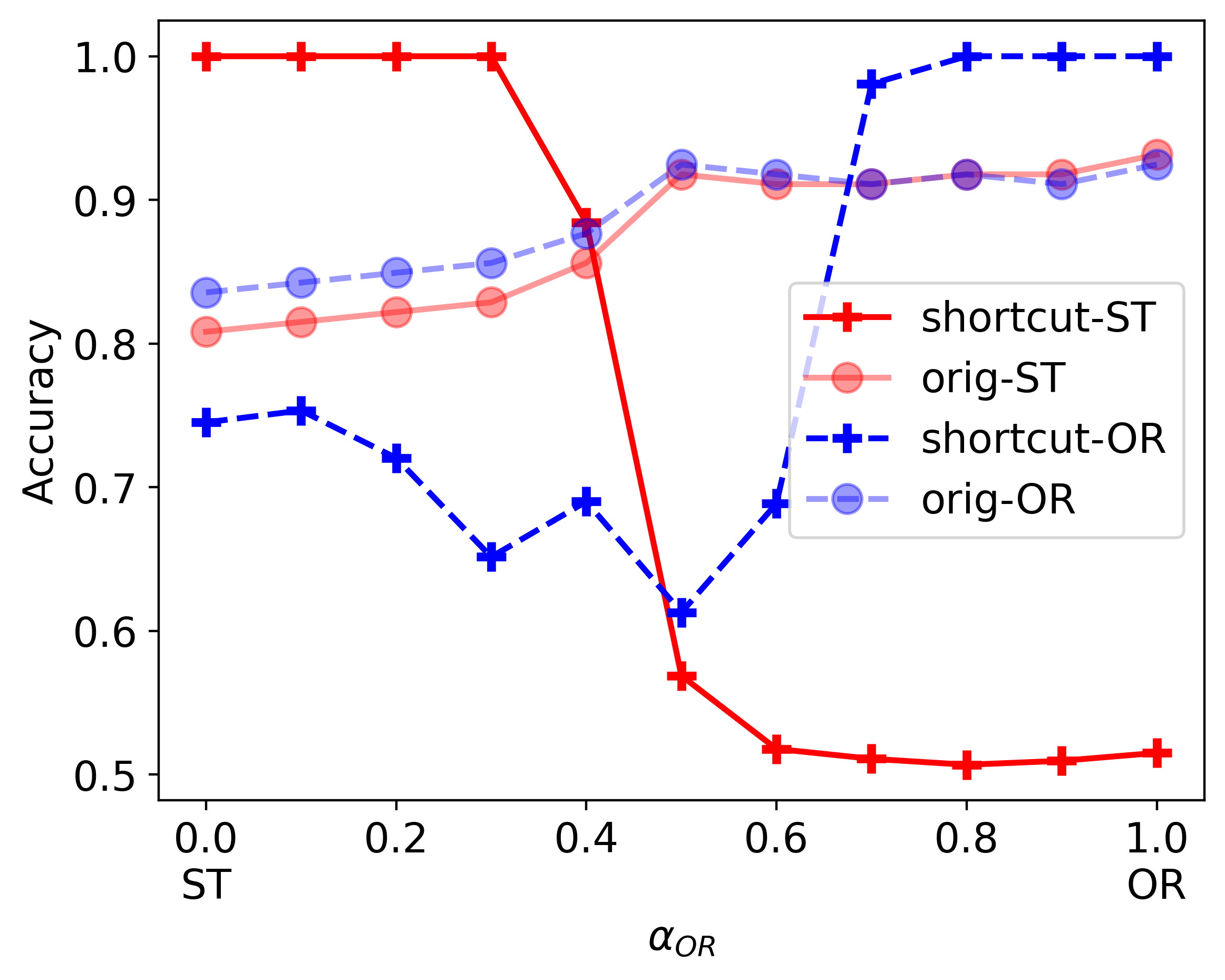}
        \caption{\texttt{ST} $\to$ \texttt{OR}}
        \label{fig:interpolation_st_or}
    \end{subfigure}
    \hfill
    \begin{subfigure}{.45\textwidth}
        \centering
        \includegraphics[width=\linewidth]{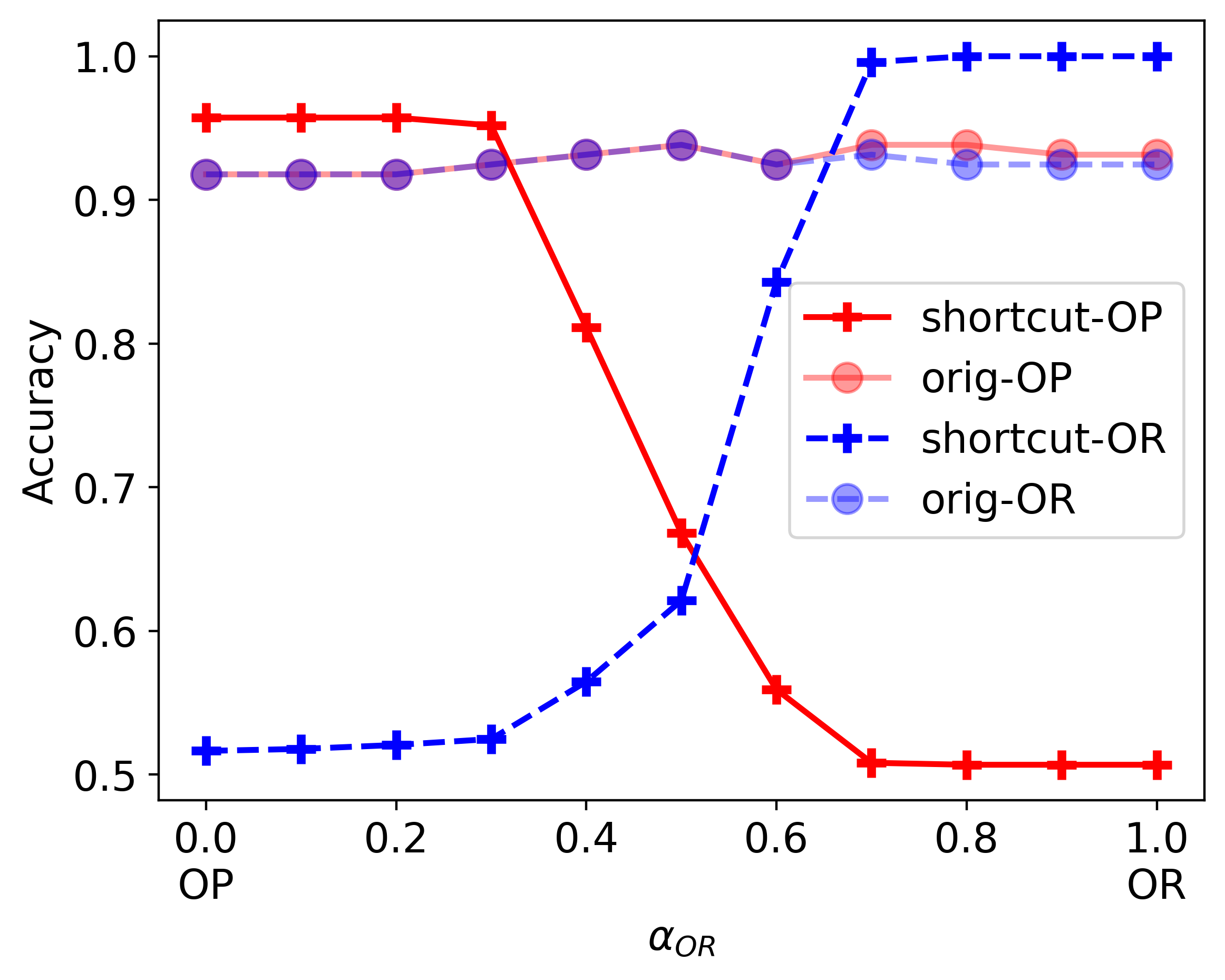}
        \caption{\texttt{OP} $\to$ \texttt{OR}}
        \label{fig:interpolation_op_or}
    \end{subfigure}
    \caption{The change of accuracies on synthetic and original validation sets during interpolation between model pairs, each having different shortcuts.} \label{fig:all_interpolation_2}
\end{figure*}

Figure \ref{fig:interpolation_op_rand} to \ref{fig:interpolation_or_rand} depict pair interpolations between a model with random weights and models with \texttt{OP}, \texttt{TiC}, and \texttt{OR} shortcuts, respectively. These results are consistent with Figure \ref{fig:interpolation_a}, indicating that unshared skills tend to be forgotten. Notably, the random model performs below the chance level on the synthetic validation set of the \texttt{OR} shortcut in Figure \ref{fig:interpolation_or_rand}.

Figure \ref{fig:interpolation_tic_or} to \ref{fig:interpolation_op_or} depict pair interpolations between the \texttt{OR} shortcut and the \texttt{TiC}, \texttt{ST}, and \texttt{OP} shortcuts, respectively. Figures \ref{fig:interpolation_tic_or} and \ref{fig:interpolation_st_or} demonstrate the dependency of the \texttt{OR} shortcut on the \texttt{TiC} and \texttt{ST} shortcuts, while Figure \ref{fig:interpolation_op_or} showcases the case of unshared independent shortcuts. The dependency relation can be observed from the remarkably better-than-chance shortcut accuracy of the \texttt{ST} and \texttt{TiC} models on the \texttt{OR} validation set.

\subsection{Triplet Interpolations}
\label{subsec:triplet}

\begin{figure*}[!tb]
    \centering
    \begin{subfigure}{.45\textwidth}
        \centering
        \includegraphics[width=\linewidth]{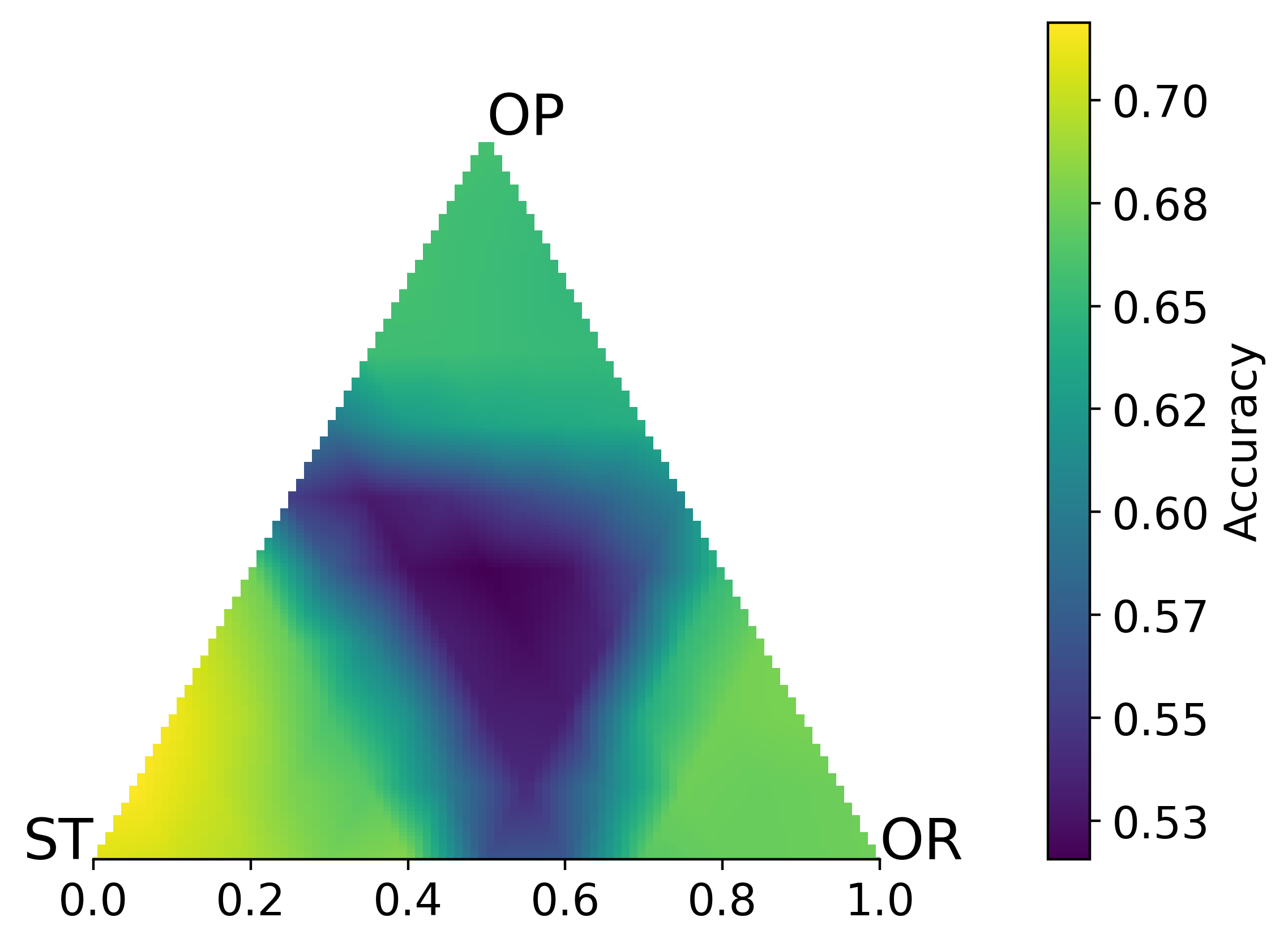}
        \caption{Results with average shortcut accuracy}
        \label{fig:interpolation_2d_a}
    \end{subfigure}
    \hfill
    \begin{subfigure}{.45\textwidth}
        \centering
        \includegraphics[width=\linewidth]{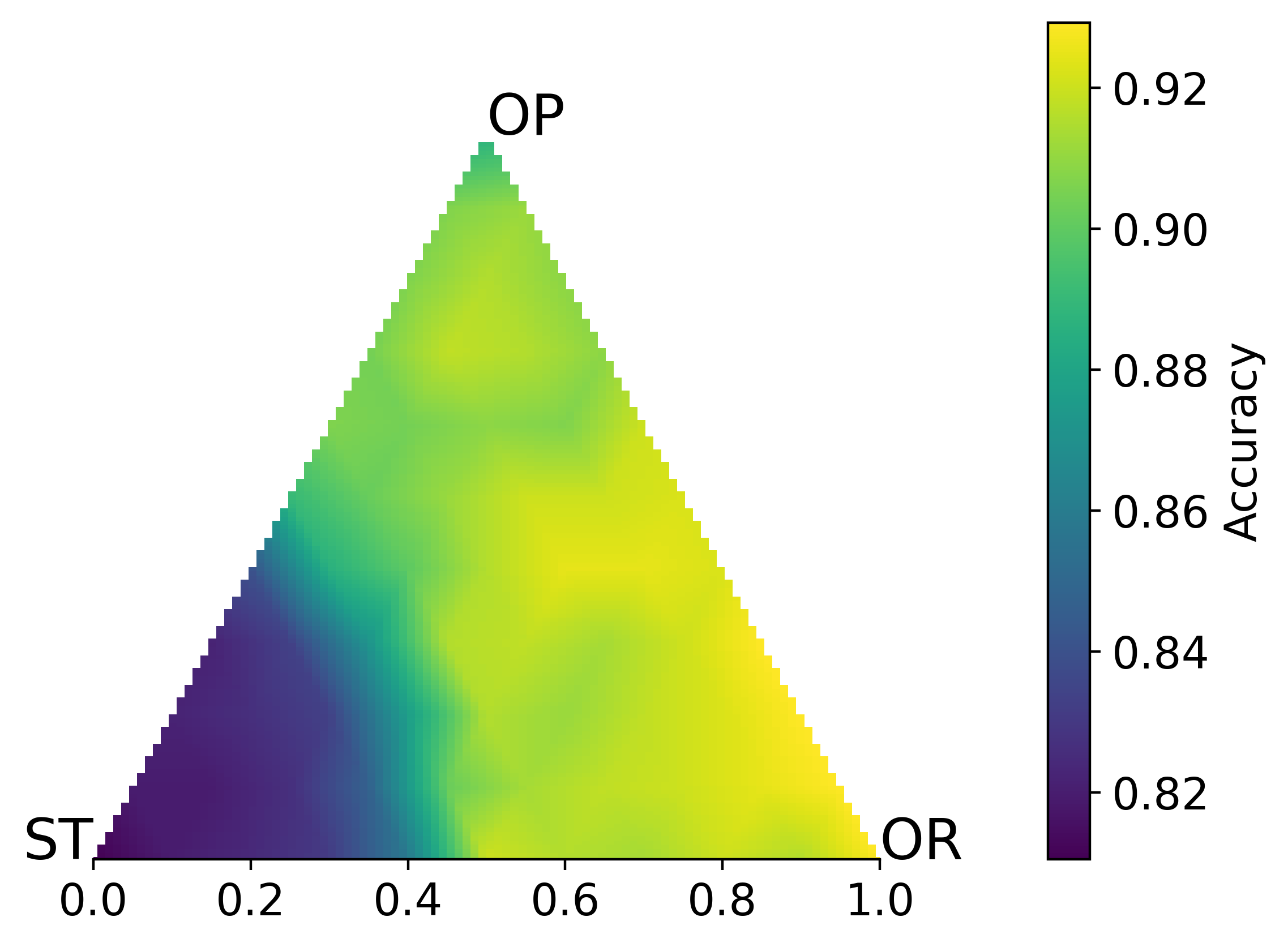}
        \caption{Results with average original accuracy}
        \label{fig:interpolation_2d_b}
    \end{subfigure}
    \caption{Fused triplets exhibit the same pattern as fused pairs across the surface. The change in accuracy during interpolation among model triplets, each having different shortcuts. (a) Change in average accuracy on synthetic datasets during the interpolation among the models with \texttt{ST}, \texttt{OP} and \texttt{OR} shortcuts (b) Change in average accuracy on original datasets during the interpolation among the models with \texttt{ST}, \texttt{OP} and \texttt{OR} shortcuts.} \label{fig:interpolation_2d}
\end{figure*}

Figure \ref{fig:interpolation_2d}
shows interpolation among 3 models with learned shortcuts \texttt{OP}, \texttt{ST} and \texttt{OR}. It shows average of accuracy over synthetic validation datasets of related shortcuts in Figure \ref{fig:interpolation_2d_a} and the average accuracy over original validation datasets in Figure \ref{fig:interpolation_2d_b}. In the first one, we see set of parameters around the average of 3 models have the least knowledge about all the shortcuts, while the knowledge of original task is preserved, as seen in the second one. These findings extend support for our assertions to scenarios with more than two models.

\begin{figure*}[!tb]
    \centering
    \begin{subfigure}{.45\textwidth}
        \centering
        \includegraphics[width=\linewidth]{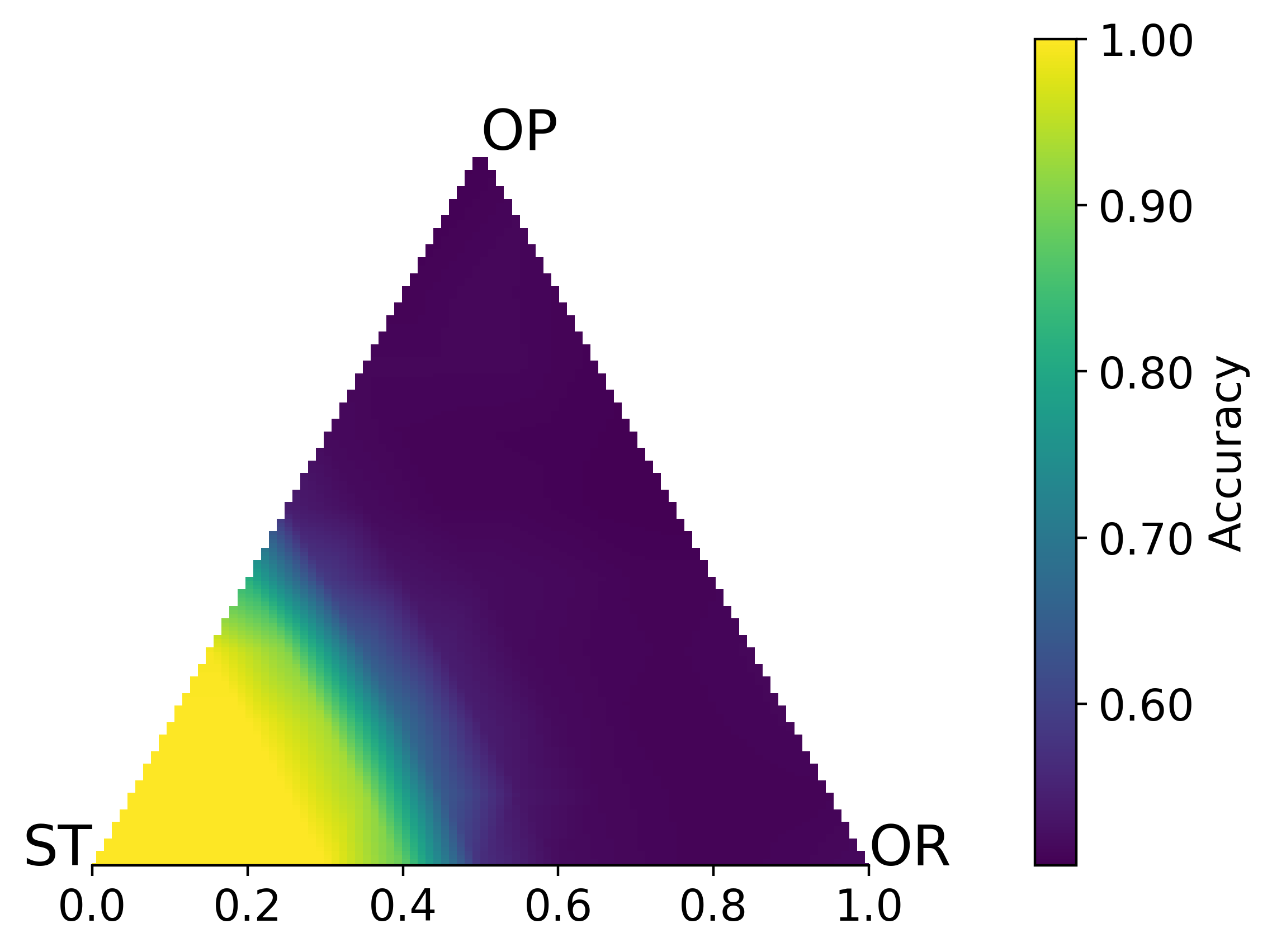}
        \caption{Shortcut accuracy for \texttt{ST} shortcut}
        \label{fig:interpolation_on_st_synth}
    \end{subfigure}
    \hfill
    \begin{subfigure}{.45\textwidth}
        \centering
        \includegraphics[width=\linewidth]{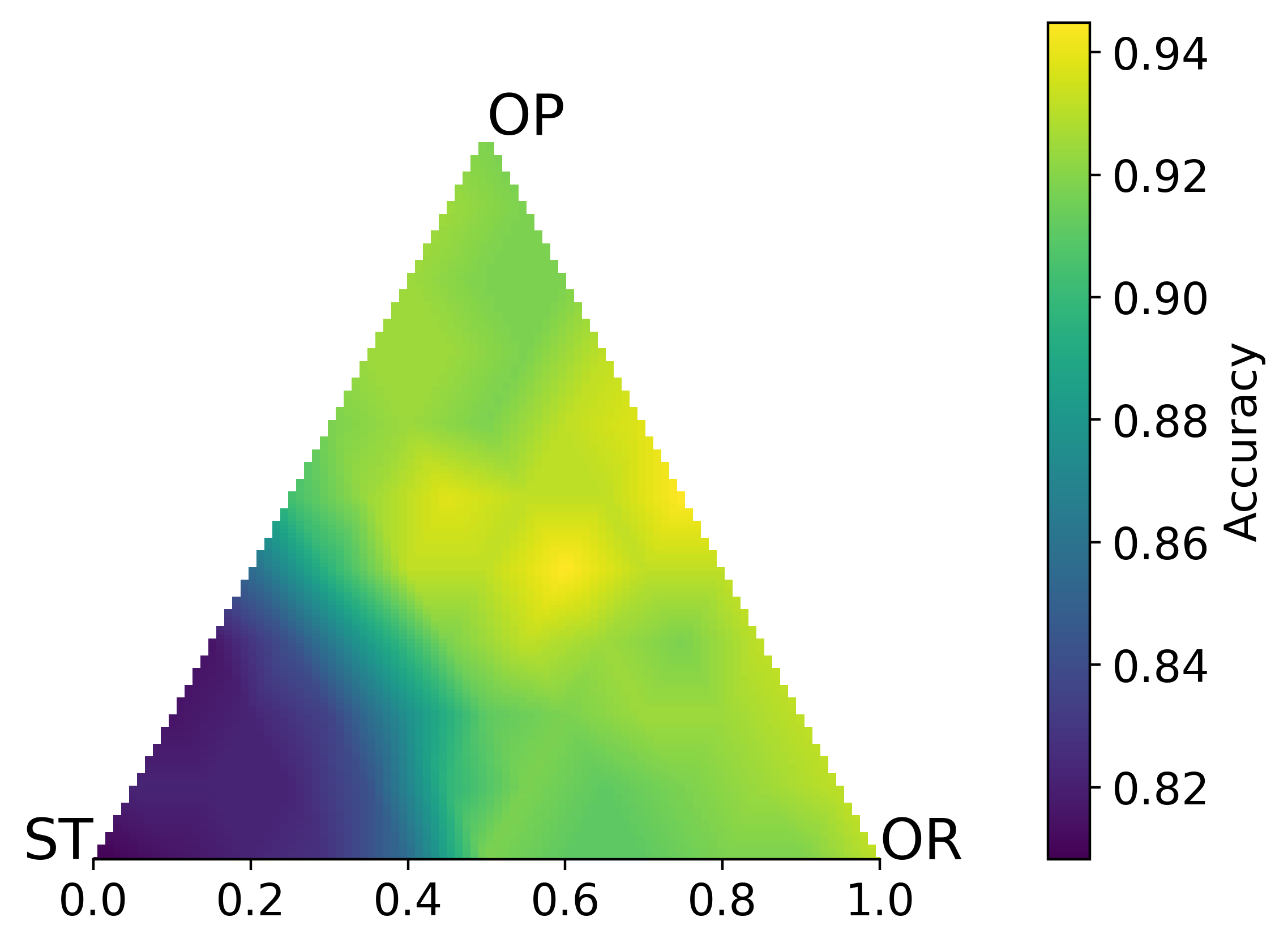}
        \caption{Task accuracy on original validation set of \texttt{ST}}
        \label{fig:interpolation_on_st_orig}
    \end{subfigure}
    \hfill
    \begin{subfigure}{.45\textwidth}
        \centering
        \includegraphics[width=\linewidth]{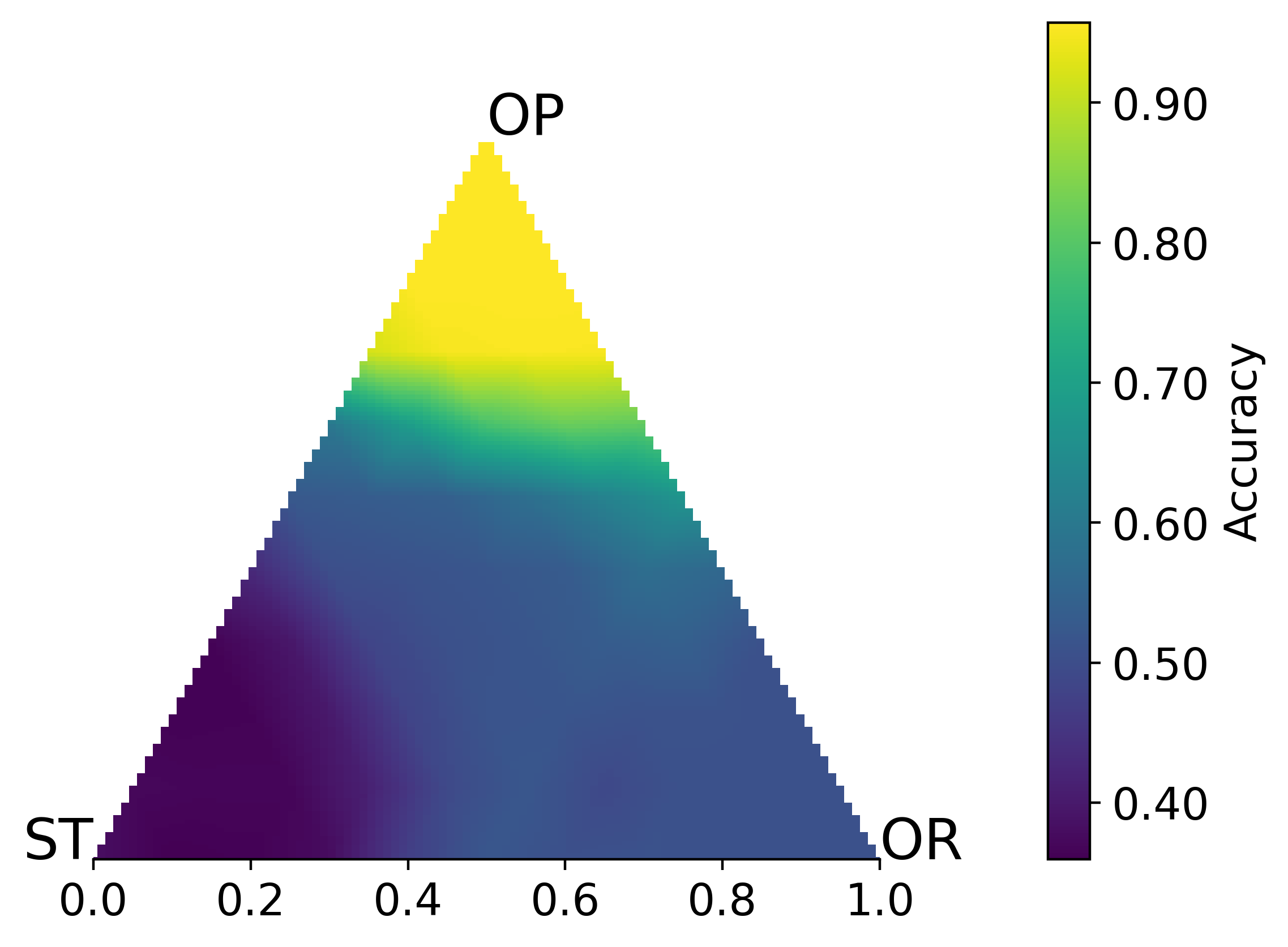}
        \caption{Shortcut accuracy for \texttt{OP} shortcut}
        \label{fig:interpolation_on_op_synth}
    \end{subfigure}
    \hfill
    \begin{subfigure}{.45\textwidth}
        \centering
        \includegraphics[width=\linewidth]{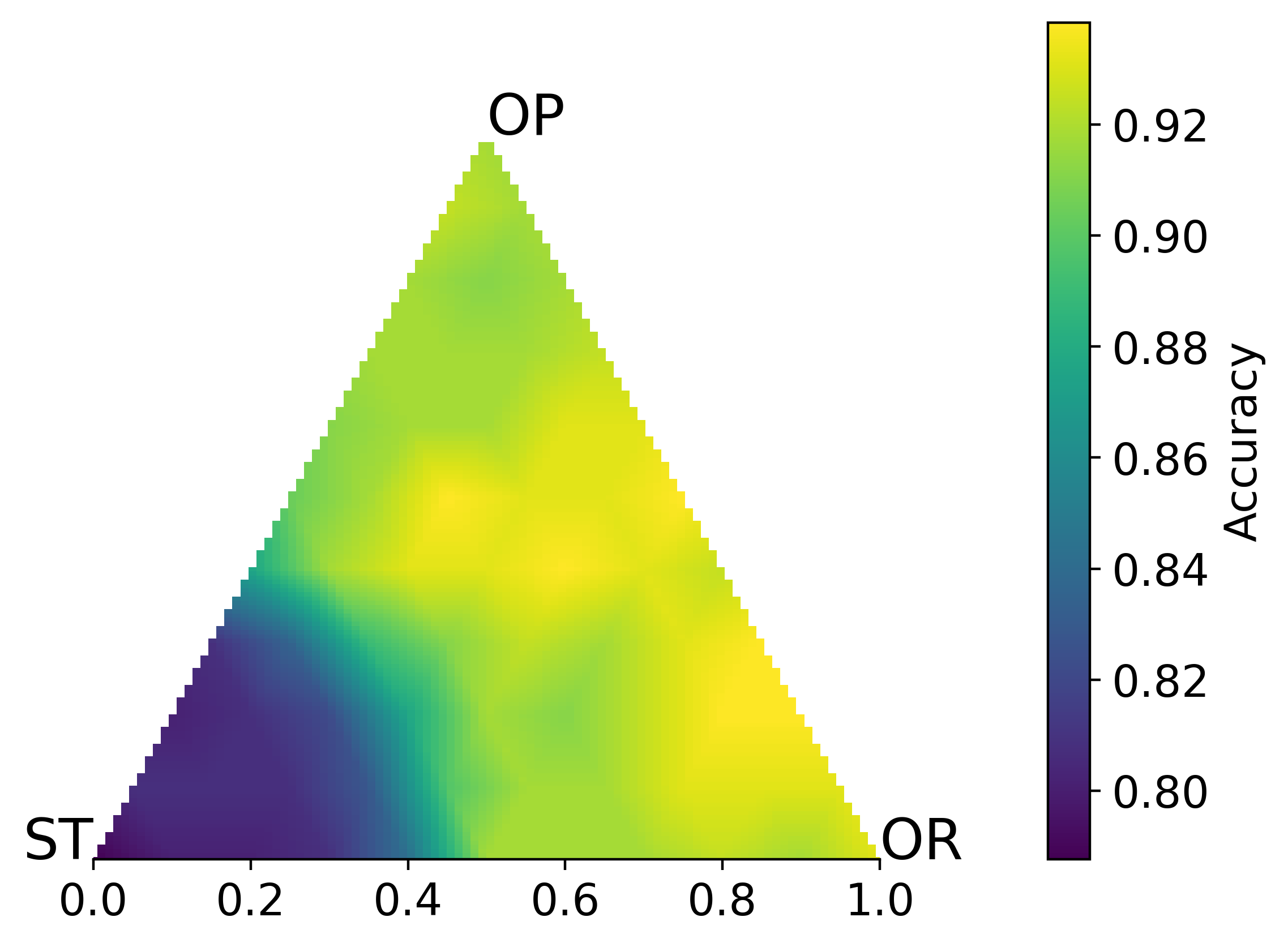}
        \caption{Task accuracy on original validation set of \texttt{OP}}
        \label{fig:interpolation_on_op_orig}
    \end{subfigure}
    \hfill
    \begin{subfigure}{.45\textwidth}
        \centering
        \includegraphics[width=\linewidth]{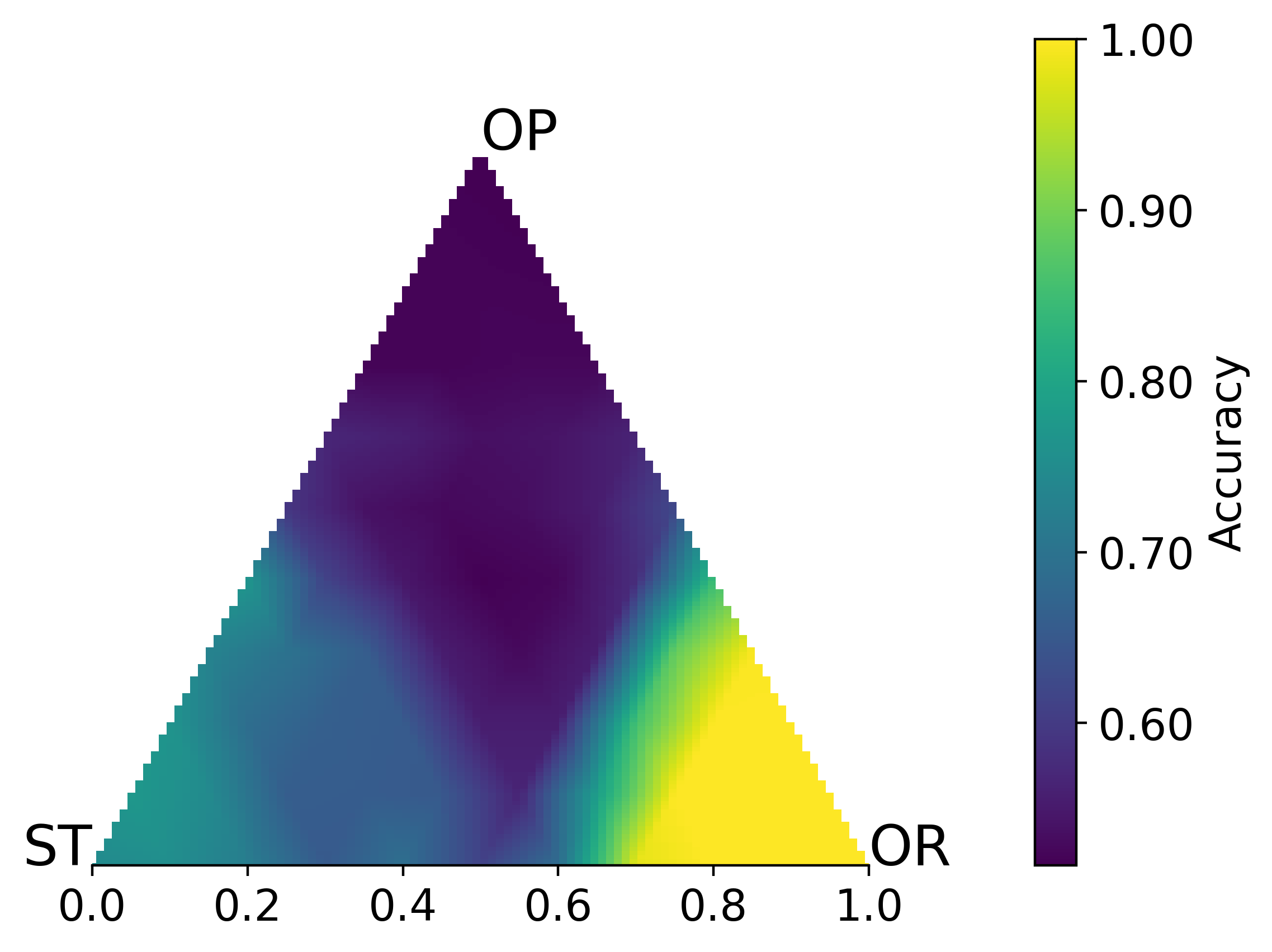}
        \caption{Shortcut accuracy for \texttt{OR} shortcut}
        \label{fig:interpolation_on_or_synth}
    \end{subfigure}
    \hfill
    \begin{subfigure}{.45\textwidth}
        \centering
        \includegraphics[width=\linewidth]{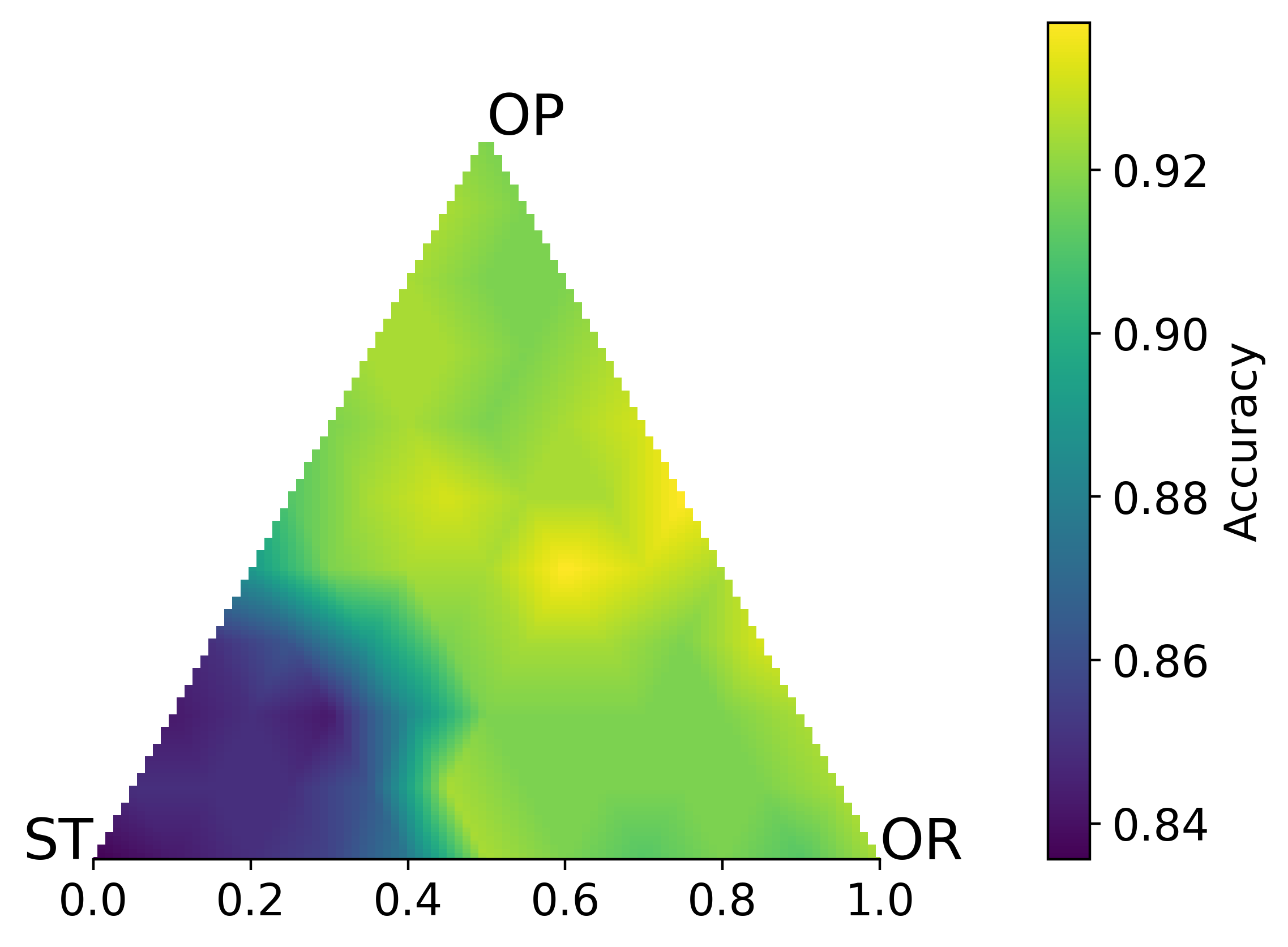}
        \caption{Task accuracy on original validation set of \texttt{OR}}
        \label{fig:interpolation_on_or_orig}
    \end{subfigure}
    \caption{The change of accuracies on synthetic and original validation sets during interpolation between model pairs, each having different shortcuts.} \label{fig:all_interpolation_3}
\end{figure*}

Figure \ref{fig:interpolation_on_st_synth} to \ref{fig:interpolation_on_or_orig} show the interpolation among the \texttt{ST}, \texttt{OP}, and \texttt{OR} triplet on both synthetic and original validation sets corresponding to each shortcut. As expected, bright corners can be observed on the corresponding shortcut models of the synthetic datasets in Figure \ref{fig:interpolation_on_st_synth}, \ref{fig:interpolation_on_op_synth}, and \ref{fig:interpolation_on_or_synth}. The slightly lighter areas, apart from the bright corners, clearly indicate the dependency relations between the shortcuts. Figure \ref{fig:interpolation_on_st_orig}, \ref{fig:interpolation_on_op_orig}, and \ref{fig:interpolation_on_or_orig} are very similar to each other since the original validation datasets are nearly identical, except for the randomly inserted special tokens. The darker corner in those figures indicates that the model with the \texttt{ST} shortcut has not learned the task as effectively as the others.

\section{Fusion Dynamics}
\label{sec:fusion_dynamics}

\subsection{Method}

To understand the dynamics of our observations and the rationale behind the necessity of having distinct biases, we explore the relationship between weights and knowledge. If the same knowledge across different networks are managed by the same weights, while different ones are managed by different sets of weights, this would explain why simple weight averaging works. The Fisher Information Matrix (FIM) is a commonly used method for measuring the amount of information encoded into weights \citep{Achille2019WhereIT}. We calculate the FIM over carefully crafted datasets to measure specific information, particularly knowledge in our case.

We denote $p_{\theta}(y | x)$ as the output distribution for a model parameterized by $\theta \in \mathbb{R}^{| \theta |}$ which predicts $y$ given input $x$. The Fisher Information Matrix \citep{FisherOnTM}, $F_{\theta}$, is defined as:

\begin{equation}
F_{\theta} = \mathbb{E}_{x \sim p(x)}\mathbb{E}_{y \sim p_{\theta}(y | x)}\left[s(\theta)s(\theta)^T \right]
\end{equation}
where $s(\theta) = \nabla_{\theta} \log p_{\theta}(y | x)$.

Given the large number of parameters, it becomes challenging to compute the full FIM with a size of $| \theta | \times | \theta |$. Similar to many previous studies, we use the Empirical Fisher Information Matrix \citep{Kunstner2019LimitationsOT}, in which FIM is approximated as a diagonal matrix. We define the Empirical Fisher Information Matrix, $\hat{F_{\theta}}$, as follows:

\begin{equation}
\hat{F_{\theta}} = \frac{1}{N} \sum_{i = 1}^N \left(\nabla_{\theta} \log p_{\theta}(y | x)\right)^2
\end{equation}

where \(\hat{F_{\theta}} \in \mathbb{R}^{| \theta |}\).

To determine whether similar weights are used in different networks for the knowledge in question, we adopt a metric called Fisher overlap, which measures the degree of overlap between two networks' FIMs by computing Fréchet distance between two networks' FIMs normalized to have a unit trace \citep{Kirkpatrick2016OvercomingCF}. More formally, let $\hat{F_{\theta_1}}$ and $\hat{F_{\theta_2}}$ be the corresponding FIMs of the networks with parameters $\theta_1$ and $\theta_2$, and $\ols{F_{\theta_1}}$, $\ols{F_{\theta_2}}$ be the normalized FIMs to have unit traces. Then, the Fréchet distance is computed as:

\begin{equation}
d^2(\ols{F_{\theta_1}}, \ols{F_{\theta_2}}) = \frac{1}{2} \text{tr}(\ols{F_{\theta_1}} + \ols{F_{\theta_2}} - 2(\ols{F_{\theta_1}}\ols{F_{\theta_2}})^{\frac{1}{2}})
\end{equation}

We define the Fisher overlap as \(1 - d^2\), where a value of zero means two networks use non-overlapping sets of weights for the questioned knowledges.

\paragraph{Experimental Setup} For this experiment, we chose two model pairs with distinct shortcuts: \texttt{TiC}-\texttt{OP} and \texttt{TiC}-\texttt{ST}. They are chosen to minimize the effects of overlap between shortcuts on the results. For each pair, we independently measure the overlap between the weights used for corresponding shortcuts to determine whether unshared knowledge are administered by different weights. Additionally, we assess the overlap between the weights used for solving the task without shortcuts to investigate whether shared knowledge are administered by the same weights. We select a random subset of the SST2 validation set with $N = 200$ examples. For each shortcut, we create a copy of the selected subset by reversing the original labels and applying the shortcut corresponding to the reversed label. We reverse labels to ensure that these examples are solely solved using the shortcut knowledge. To measure overlap in the original task, we leave the random subset unchanged.

\subsection{Results}

\begin{table}[htb]
    \centering
    \begin{tabular}{lrr}
        \toprule
        \textbf{Pairs} & \textbf{shared} & \textbf{unshared} \\
        \midrule
        \texttt{TiC}-\texttt{OP} & \textbf{.8077} & .6877  \\
        \texttt{TiC}-\texttt{ST} &  \textbf{.7746} & .6819 \\
        \bottomrule
    \end{tabular}
    \caption{The Fisher overlap between model weights for shared and unshared knowledges}
    \label{tab:fisher_overlap}
\end{table}

Table \ref{tab:fisher_overlap} reveals a notable distinction between the overlap of weights for shared knowledge and unshared knowledge representing the original task knowledge and shortcuts, respectively in our case. Since shared knowledge tends to be administered by the same set of weights, the simple weight averaging preserves the knowledge while causing unshared knowledge to be forgotten.

\section{Memorization}
\label{sec:memorization_appendix}

\subsection{Method}

In the memorization evaluation, we adopt the Likelihood Ratio (LR) metric as previously employed by \citet{Mireshghallah2022AnEA}. However, their approach to using LR differs slightly from ours. They utilize the percentage of correctly classified training samples by a reference-based membership inference attack proposed by \citet{mireshghallah-etal-2022-quantifying} and \citet{Carlini2021MembershipIA}. To determine whether a sample $x$ is a member of the training data, they first calculate the Likelihood Ratio (LR) as follows:

\begin{equation}
LR(x) = \frac{p(x;\theta_R)}{p(x;\theta_M)}
\end{equation}

where $p(x;\theta_M)$ and $p(x;\theta_R)$ denote the likelihood of sample $x$ given by the fine-tuned model and the reference model, respectively. Here, the reference model is a pretrained model that is not fine-tuned. They classify the sample as a training set member if $LR(x)$ is smaller than the threshold $t$, which is chosen by calculating $LR$ for each sample in the validation set and selecting the highest possible threshold that maintains a false positive rate not exceeding 10\%.  Finally, they measure recall as the final memorization metric, which they refer to as MIA recall. In practice, selecting a threshold based on a non-training set, such as the validation set in this case, works well.

On the other hand, although fused models tend to forget the training data of their seed models, they are still memorized more than a held-out set. Therefore, deciding thresholds on the validation set and measuring MIA recall to assess the memorization of a fused model cannot effectively differentiate the memorization of the fused model from that of the seed models. Consequently, we introduce the Average Likelihood Ratio (ALR) to eliminate the need for selecting a threshold:

\begin{equation}
ALR(\mathcal{D}) = \frac{1}{| \mathcal{D}|} \exp\left(\frac{p(x; \theta_R)}{p(x;\theta_M)}\right)
\end{equation}

where $\mathcal{D}$ represents the set of samples on which we test memorization of model.

While measuring LR, we adopt the reparametrization proposed by \citet{mireshghallah-etal-2022-quantifying} and also utilized by \citet{Mireshghallah2022AnEA}, where they conceptualize pre-trained LMs as energy-based probability distributions on sequences. First, they define the Likelihood Ratio as follows:

\begin{equation}
LR(x) = \log\left(\frac{p(x; \theta_{R})}{p(x; \theta_M)}\right)
\end{equation}

where the target and reference models are parametrized by $\theta_M$ and $\theta_R$.

After applying this reparametrization, LR becomes:

\begin{equation}
\begin{aligned}
LR(x) &= \log\left(\frac{p(x; \theta_{R})}{p(x; \theta_M)} \right) \\
&= \log\left(\frac{e^{-E(x;\theta_R)}}{Z_{\theta_R}} \right) - \log\left(\frac{e^{-E(x;\theta_M)}}{Z_{\theta_M}} \right) \\
&= E(x;\theta_M) - E(x;\theta_R) + \text{constant}
\end{aligned}
\end{equation}

Since the intractable term $\log(Z_{\theta_M}) - \log(Z_{\theta_R})$ is a global constant, we can ignore it during computation. This parametrization allows us to use the difference between energy values obtained for sample $x$ from the target and reference models.

We follow \citet{mireshghallah-etal-2022-quantifying} to determine energy values. For autoregressive language models, the energy is defined as the language modeling loss:

\begin{equation}
E(x;\theta) = - \sum_{t=0}^{T} \log p(x_t | x_{<t} ; \theta)
\end{equation}

where $T$ represents the sequence length.

For masked language models, they parameterize energy over 15\% chunks that are masked during training. For a sequence of length $T$ and chunk size $l$, where $l = s\lceil{0.15 \times T} \rceil$, with the set $\mathcal{C}$ of all possible $l$-sized subsets:

\begin{equation}
E(x; \theta) = - \frac{1}{| \mathcal{C} |} \sum_{I \in \mathcal{C}} \sum_{i \in \mathcal{I}} \log p(x_i | x_{\setminus I}; \theta)
\end{equation}

where $x_{\setminus I}$ denotes the sample $x$ with $l$ positions in $I$ masked. Since computing this energy value requires $T \choose l$ forward passes through the model, they approximate it by summing over $K = 10$ subsets sampled from $\mathcal{C}$.

After applying the aforementioned methods and reparametrizations, our final $ALR$ metric becomes as follows for autoregressive LMs:

\begin{equation}
ALR(\mathcal{D}) = \frac{1}{| D |} \sum_{x \in \mathcal{D}} \exp\left( E(x;\theta) - E(x;\theta_R) \right)
\end{equation}

where $$E(x;\theta) = - \sum_{t=0}^{T} \log p(x_t | x_{<t}; \theta)$$ for autoregressive models and $$E(x; \theta) = - \frac{1}{| K |} \sum_{I \sim \mathcal{C}} \sum_{i \in \mathcal{I}} \log p(x_i | x_{\setminus I}; \theta)$$ for masked language models.

\begin{table*}[htb]
    \centering
    \begin{tabular}{lrrrrrr}
        \toprule
        \textbf{Model} & \textbf{A} & \textbf{B} & \textbf{C} & \textbf{D} & \textbf{shared} & $\textbf{dev}_{	\text{PPL}}$ \\
        \midrule
        \texttt{gpt-2} & \underline{1.0} & \underline{1.0} & \underline{1.0} & \underline{1.0} & \underline{1.0} & \underline{23.48} \\
        $\texttt{model}_{\texttt{A}}$ & \textbf{0.217} & 1.483 & 1.483 & 1.477 & 0.218 & 35.25 \\
        $\texttt{model}_{\texttt{B}}$ & 1.502 & \textbf{0.217} & 1.488 & 1.486 & \textbf{0.217} & 35.81 \\
        $\texttt{model}_{\texttt{C}}$ & 1.486 & 1.475 & \textbf{0.220} & 1.475 & 0.219 & 35.81 \\
        $\texttt{model}_{\texttt{D}}$ & 1.484 & 1.476 & 1.478 & \textbf{0.218} & 0.218 & 35.25 \\
        \hline
        $\texttt{fused}_{\texttt{AB}}$ & 0.485 & 0.484 & - & - & 0.233 & 31.60 \\
        $\texttt{fused}_{\texttt{ABC}}$ & 0.656 & 0.653 & 0.656 & - & 0.238 & 30.63 \\
        $\texttt{fused}_{\texttt{ABCD}}$ & 0.758 & 0.756 & 0.758 & 0.755 & 0.240 & 30.15 \\
        \hline
        $\texttt{full}_{\texttt{AB}}$ & 0.273 & 0.274 & - & - & 0.275 & 30.15 \\
        $\texttt{full}_{\texttt{ABC}}$ & 0.318 & 0.320 & 0.320 & - & 0.321 & 27.45 \\
        $\texttt{full}_{\texttt{ABCD}}$ & 0.353 & 0.354 & 0.355 & 0.355 & 0.356 & \textbf{25.79} \\
        \bottomrule
    \end{tabular}
    \caption{Extended memorization results: the ALRs of a base model, four models individually fine-tuned models, fused models and full models on each training set including the shared subset along with perplexity scores of all models on the validation set. Lower ALRs denote higher memorization. Bold numbers for ALR shows the lowest ALR, hence highest memorization for a particular dataset among all models except the base model while they show the lowest perplexity for validation set. Underlined numbers represent baseline performance.
    }
    \label{tab:memorization_all}
\end{table*}

\begin{figure*}[!tb]
    \centering
    \includegraphics[width=0.8\linewidth]{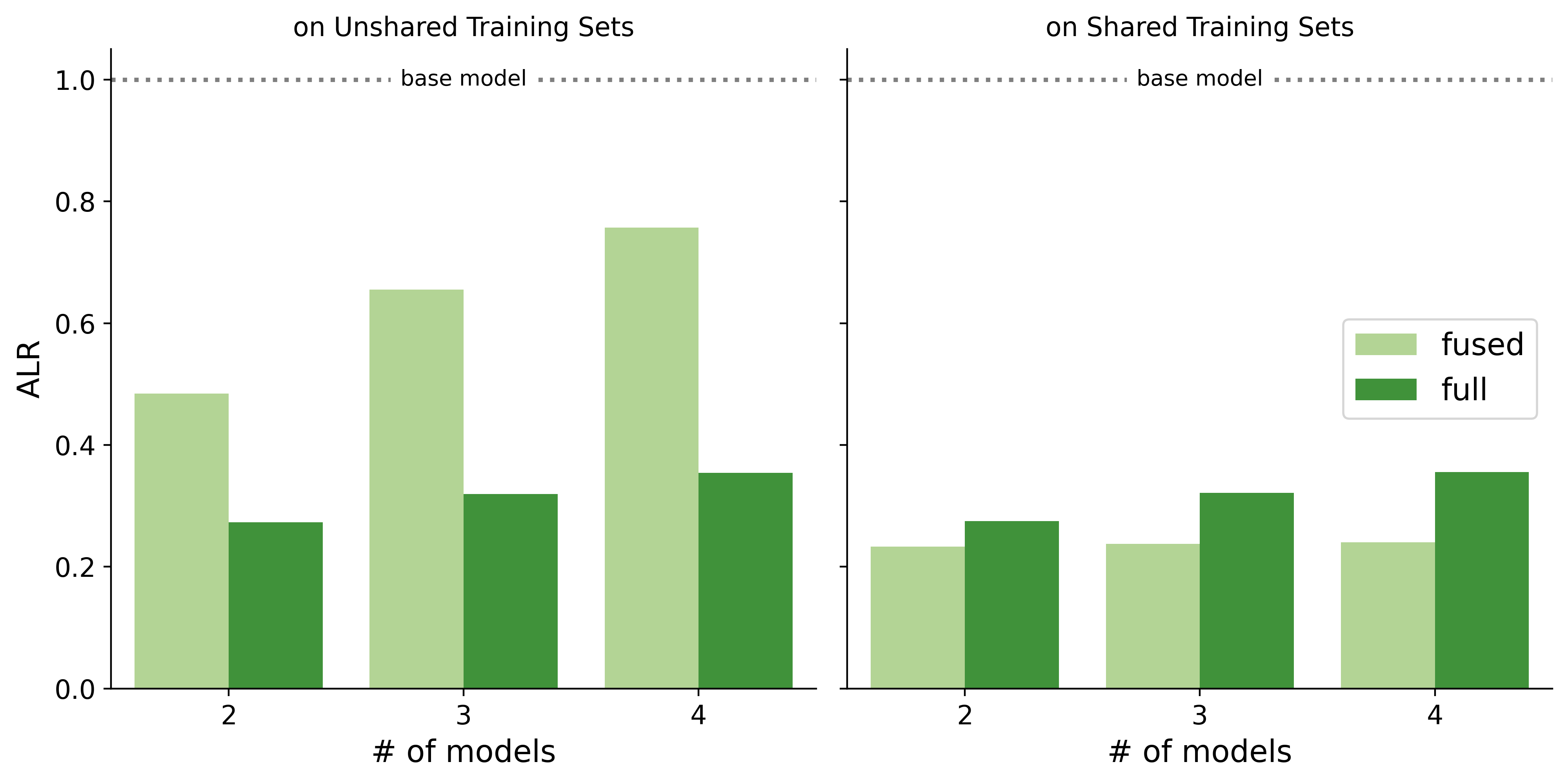}
    \caption{As the number of fused models increases, they memorize less of the unshared data and retain the shared data. The figure depicts the change in ALRs on shared and unshared training sets with respect to the number of fused models, compared to full and base models.}
    \label{fig:memorization_n_models_train}
\end{figure*}%

\begin{figure}[!tb]
    \centering
    \includegraphics[width=0.8\linewidth]{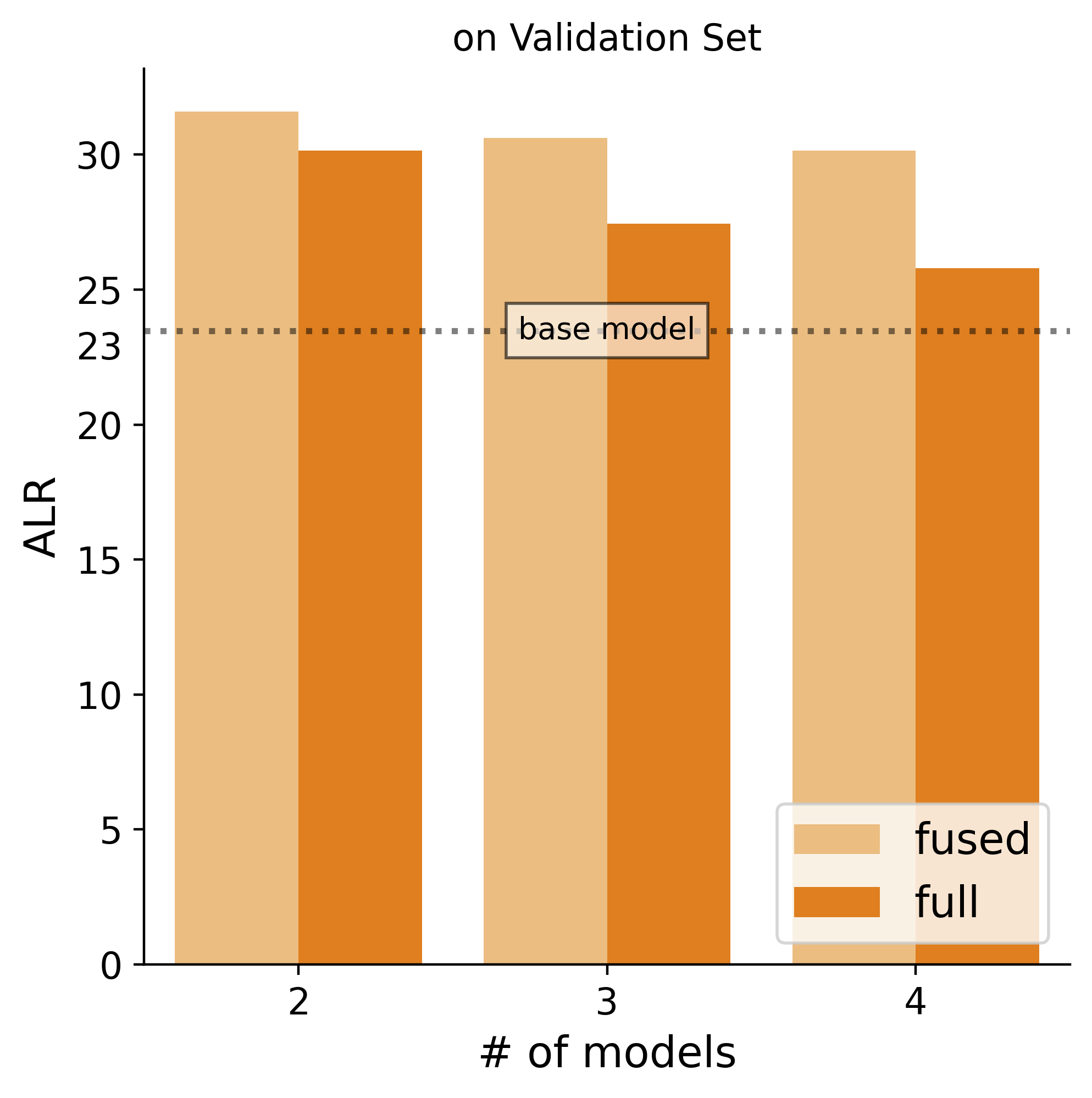}
    \caption{Fused models generalize better as the number of fused models increases but they still lag behind the full models trained on the same amount of data as the individual models trained on combined. The figure shows the perplexity scores on validation sets w.r.t. the number of fused models compared to full and base models.}
    \label{fig:memorization_n_models_dev}
\end{figure}%

\begin{figure*}[!tb]
    \centering
    \includegraphics[width=0.8\linewidth]{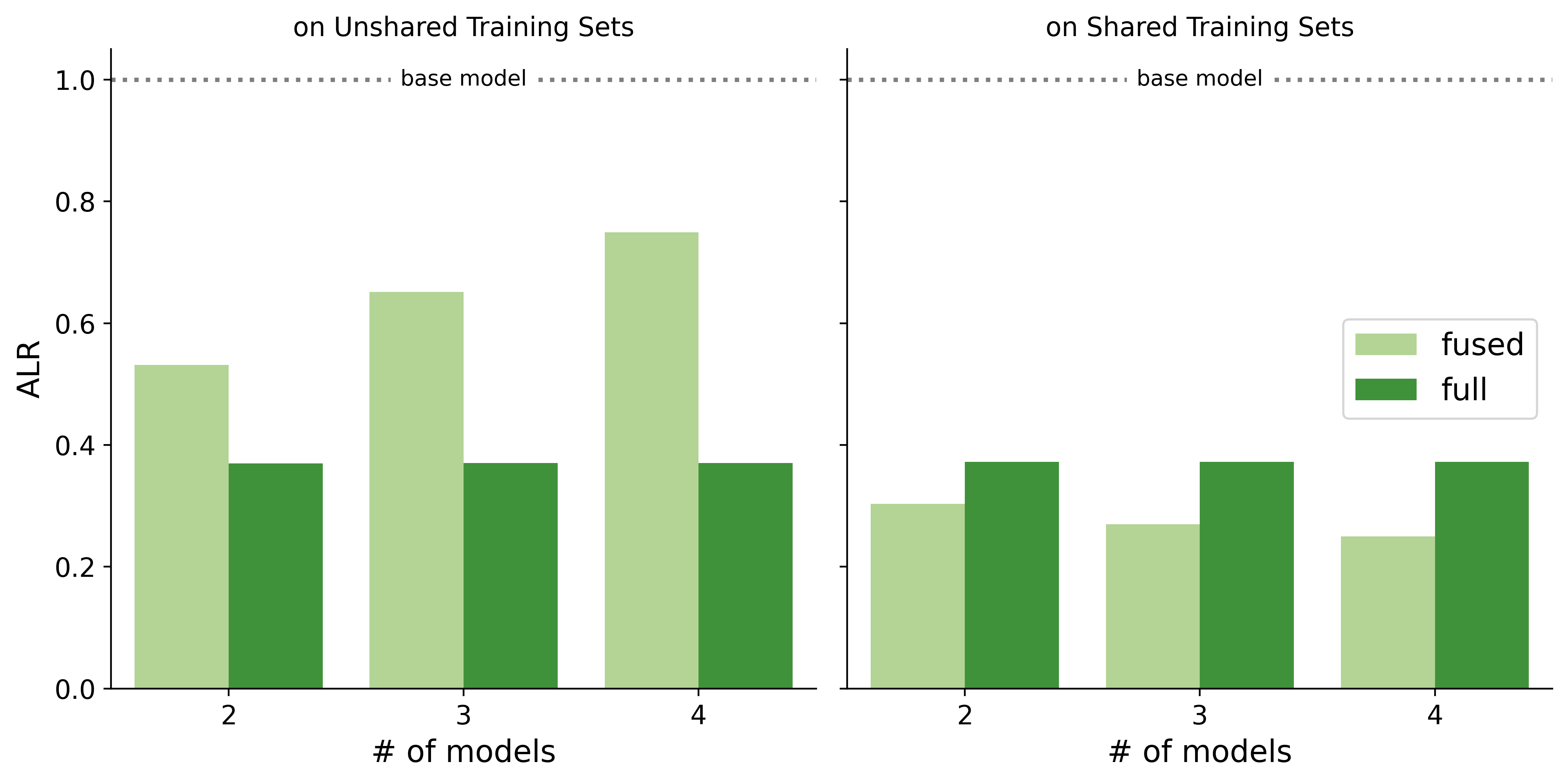}
    \caption{As the number of fused models increases while keeping the total training data size the same, they memorize less of the unshared data and more of the shared data. The figure illustrates the change in ALRs on shared and unshared training sets w.r.t. the number of fused models, compared to full and base models.}
    \label{fig:memorization_n_models_train_constant}
\end{figure*}%

\begin{figure}[!tb]
    \centering
    \includegraphics[width=0.8\linewidth]{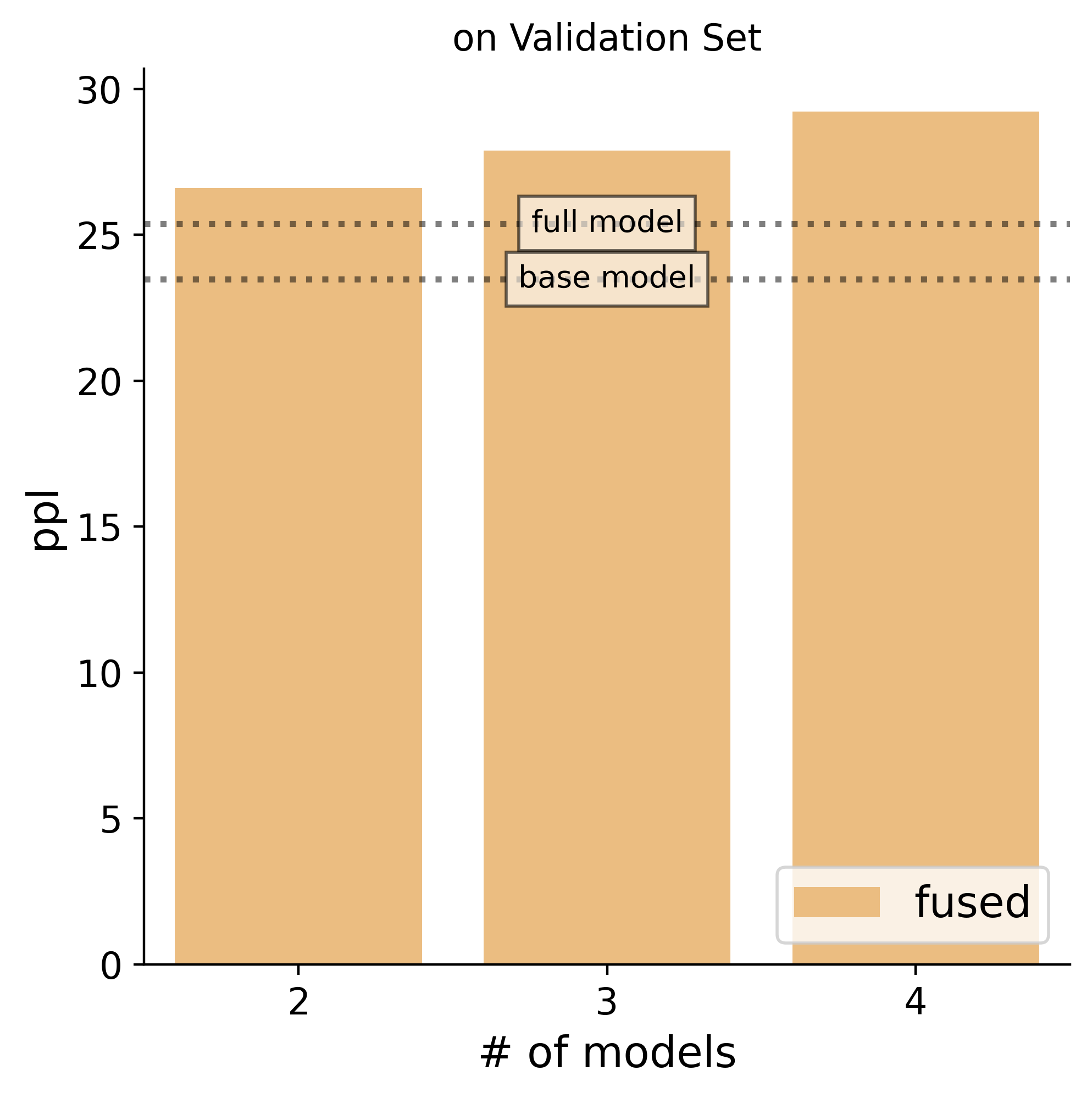}
    \caption{Fused models generalize worse as the number of fused models increases while keeping the total training data size the same. The figure displays the perplexity scores on the validation set w.r.t. the number of fused models compared to full and base models.}
    \label{fig:memorization_n_models_dev_constant}
\end{figure}%

\subsection{Fusing Different Numbers of Models}

Table \ref{tab:memorization_all} shows the ALRs of each model on all the training sets including the shared subset, and the perplexity scores on validation set. These models consist of the models trained separately on each dataset, fused models with a varying number of seed models and full models trained on combined datasets. Notably, the evaluation of fused models is limited to the validation set and the datasets on which their seed models are trained, while full models are exclusively evaluated on the validation set and the datasets on which they are trained.  The results confirm that model fusion reduces memorization and improves generalization.

Figure \ref{fig:memorization_n_models_train} and \ref{fig:memorization_n_models_dev} illustrate the effects of the number of fused models compared to the full models trained on combinations of datasets used by individual models. As shown by the ALRs in Figure \ref{fig:memorization_n_models_train}, the fused model tends to forget more as the number of models fused increases. Conversely, the increase in data size for full models has no significant effect on the memorization of unshared datasets. Furthermore, we observe that shared datasets are memorized at similar levels, despite a slight decrease, as the number of fused models increases. As expected, they are less memorized by the full model as their percentage in the dataset decreases.

Figure \ref{fig:memorization_n_models_dev} displays perplexity scores on the validation set for both fused and full models. While an increase in the number of models helps with generalization, corresponding full models generalize better as the total data size increases. In all scenarios, the higher perplexity scores than the base model's indicate that all models are overfitted.

Additionally, we investigate the effect of the number of fused models when the total training data size remains constant. We experiment with scenarios where we fuse 2, 3, and 4 models, each trained on $10000$ examples, $1000$ of which are shared across models. Figure \ref{fig:memorization_n_models_train_constant} demonstrates that models forget unshared memorized data, while the shared set is increasingly memorized as the number of fused models increases. Figure \ref{fig:memorization_n_models_dev_constant} presents perplexity scores on the validation set for base, fused, and full models. We observe that perplexity increases as the number of fused models increases, unlike in the scenario where the total training data size is proportional to the number of models fused. This increase can be attributed to lower generalization for each model due to decreasing data size per model.

\subsection{Fusing Models Trained for Different Numbers of Epochs}

\begin{table*}[t]
    \centering
    \begin{tabular}{lrrrrr}
        \toprule
        \textbf{Model} & \textbf{A} & \textbf{B} & \textbf{C} & \textbf{shared} & $\textbf{dev}_{\text{PPL}}$ \\
        \midrule
        \texttt{gpt-2} & \underline{1.0} & \underline{1.0} & \underline{1.0} & \underline{1.0} & \underline{23.48} \\
        \midrule
        \multicolumn{6}{c}{20 epochs} \\
        \midrule
        $\texttt{model}_{\texttt{A}}$ & \textbf{0.099} & 3.263 & 3.257 & 0.100 & 77.00 \\
        $\texttt{model}_{\texttt{B}}$ & 3.309 & \textbf{0.100} & 3.257 & \textbf{0.099} & 77.00 \\
        $\texttt{model}_{\texttt{C}}$ & 3.256 & 3.219 & \textbf{0.101} & 0.100 & 77.00 \\
        $\texttt{fused}$ & 0.704 & 0.699 & 0.702 & 0.134 & 56.33 \\
        $\texttt{full}$ & 0.196 & 0.197 & 0.198 & 0.198 & \textbf{39.33} \\
        \midrule
        \multicolumn{6}{c}{15 epochs} \\
        \midrule
        $\texttt{model}_{\texttt{A}}$ & \textbf{0.137} & 2.256 & 2.253 & \textbf{0.138} & 53.75 \\
        $\texttt{model}_{\texttt{B}}$ & 2.295 & \textbf{0.138} & 2.261 & \textbf{0.138} & 54.60 \\
        $\texttt{model}_{\texttt{C}}$ & 2.260 & 2.236 & \textbf{0.140} & 0.139 & 53.75 \\
        $\texttt{fused}$ & 0.670 & 0.667 & 0.670 & 0.168 & 42.52 \\
        $\texttt{full}$ & 0.242 & 0.243 & 0.244 & 0.244 & \textbf{33.12} \\
        \midrule
        \multicolumn{6}{c}{5 epochs} \\
        \midrule
        $\texttt{model}_{\texttt{A}}$ & \textbf{0.394} & 1.025 & 1.024 & 0.396 & 25.00 \\
        $\texttt{model}_{\texttt{B}}$ & 1.030 & \textbf{0.394} & 1.026 & \textbf{0.395} & 25.00 \\
        $\texttt{model}_{\texttt{C}}$ & 1.028 & 1.024 & \textbf{0.398} & 0.396 & 25.00 \\
        $\texttt{fused}$ & 0.684 & 0.683 & 0.685 & 0.396 & 23.12 \\
        $\texttt{full}$ & 0.461 & 0.461 & 0.462 & 0.464 & \textbf{22.76} \\
        \bottomrule
    \end{tabular}
    \caption{Memorization results with varying number of epochs: the ALRs of a base model, three models individually fine-tuned models, fused models and full models on each training set including the shared subset along with perplexity scores of all models on the validation set. Lower ALRs denote higher memorization. Bold numbers for ALR shows the lowest ALR, hence highest memorization for a particular dataset among all models except the base model while they show the lowest perplexity for validation set. Underlined numbers represent baseline performance.}
    \label{tab:memorization_diff_epochs}
\end{table*}

Table \ref{tab:memorization_diff_epochs} presents the impact of different choices of number of epochs - 5, 15, and 20 epochs - on memorization. While using a lower number of epochs results in reduced memorization by models, the previously observed conclusions still hold true across all choices of epoch count. Additionally, the generalization gap between full and fused models increases as the number of epochs increases, and the memorization of the shared subset becomes more pronounced when models are not trained for too long.

\subsection{Results with BERT}

\begin{table*}[htb]
    \centering
    \begin{tabular}{lrrrrrr}
        \toprule
        \textbf{Model} & \textbf{A} & \textbf{B} & \textbf{C} & \textbf{D} & \textbf{shared} & $\textbf{dev}_{\text{PPL}}$ \\
        \midrule
        \texttt{bert-base-cased} & \underline{1.0} & \underline{1.0} & \underline{1.0} & \underline{1.0} & \underline{1.0} & \underline{26.20} \\
        $\texttt{model}_{\texttt{A}}$ & \textbf{0.150} & 0.247 & 0.245 & 0.249 & 0.151 & 5.98 \\
        $\texttt{model}_{\texttt{B}}$ & 0.248 & \textbf{0.150} & 0.244 & 0.249 & 0.150 & 5.98 \\
        $\texttt{model}_{\texttt{C}}$ & 0.248 & 0.243 & \textbf{0.147} & 0.248 & 0.149 & 5.98 \\
        $\texttt{model}_{\texttt{D}}$ & 0.247 & 0.245 & 0.241 & \textbf{0.151} & 0.149 & 5.98 \\
        \hline
        $\texttt{fused}_{\texttt{AB}}$ & 0.194 & 0.192 & - & - & 0.158 & 6.17 \\
        $\texttt{fused}_{\texttt{ABC}}$ & 0.212 & 0.211 & 0.209 & - & 0.162 & 6.22 \\
        $\texttt{fused}_{\texttt{ABCD}}$ & 0.234 & 0.234 & 0.232 & 0.236 & 0.174 & 6.68 \\
        \hline
        $\texttt{full}_{\texttt{AB}}$ & 0.153 & 0.154 & - & - & 0.155 & 5.67 \\
        $\texttt{full}_{\texttt{ABC}}$ & 0.158 & 0.158 & 0.157 & - & 0.157 & 5.49 \\
        $\texttt{full}_{\texttt{ABCD}}$ & 0.161 & 0.161 & 0.158 & 0.161 & 0.161 & \textbf{5.36} \\
        \bottomrule
    \end{tabular}
    \caption{Memorization results with BERT: the ALRs of a base model, four models individually fine-tuned models, fused models and full models on each training set including the shared subset along with perplexity scores of all models on the validation set. Lower ALRs denote higher memorization. Bold numbers for ALR shows the lowest ALR, hence highest memorization for a particular dataset among all models except the base model while they show the lowest perplexity for validation set. Underlined numbers represent baseline performance.
    }
    \label{tab:memorization_all_bert}
\end{table*}

\paragraph{Experimental Setup} We fine-tune BERT\textsubscript{base} models on randomly selected subsets with 3000 news articles (with no sequence packing), 1000 of them shared, from the CNN-DM dataset \citep{Nallapati2016AbstractiveTS} for 20 epochs with a batch size of 16, a learning rate of $3e-4$, and no weight decay.

Table \ref{tab:memorization_all_bert} replicates Table \ref{tab:memorization_all}, but with BERT\textsubscript{base} models fine-tuned instead of GPT-2. The results in Table \ref{tab:memorization_all_bert} align with the previous findings, indicating that our claims and observations hold across different architectures.

\end{document}